\ifdefined\XeTeXversion\else\pdfoutput=1\fi
\documentclass[10pt,letterpaper]{article}

\usepackage[utf8]{inputenc}
\usepackage[T1]{fontenc}
\usepackage{newtxtext}

\usepackage{amsmath,amssymb,amsthm}
\usepackage[varvw]{newtxmath}
\usepackage[letterpaper,margin=1in]{geometry}
\usepackage{microtype}
\usepackage{graphicx}
\usepackage{booktabs}
\usepackage{adjustbox}
\usepackage{longtable}
\usepackage{pdflscape}
\usepackage{float}
\usepackage{array}
\usepackage{enumitem}
\usepackage[font=small]{caption}
\setlist{nosep,leftmargin=*}
\usepackage[round,semicolon,authoryear]{natbib}
\usepackage{url}
\usepackage{xcolor}
\usepackage{tikz}
\usetikzlibrary{positioning,arrows.meta}
\usepackage{hyperref}

\newcommand{\papertitle}{Learning a calibrated decision head from logged bandit feedback}
\newcommand{\paperkeywords}{logged bandit feedback, calibration, proper scoring rules, decision heads, selective classification, LoRA}

\hypersetup{
  colorlinks=true,
  linkcolor={blue!60!black},
  citecolor={green!40!black},
  urlcolor={blue!70!black},
  pdftitle={\papertitle},
  pdfauthor={Ye Yuan},
  pdfsubject={Machine learning},
  pdfkeywords={\paperkeywords}
}

\theoremstyle{plain}
\newtheorem{corollary}{Corollary}

\title{\papertitle}
\author{Ye Yuan\thanks{Ph.D. in Physics, University College Dublin.}\\
  Independent Researcher\\
  \texttt{ye.yuan.phd@gmail.com}}
\date{}

\begin{document}
\maketitle
\begin{abstract}
A decision head reads a state and a list of options and returns a probability per option in one forward pass. Once deployed, it usually observes only whether the option it chose was right: logged bandit feedback. We adapt such a head, LoRA on Qwen3-0.6B-Base and 1.7B-Base, from that feedback and report three results. First, the Brier or log score of the sampled action's own probability is not a usable REINFORCE reward: every wrong action earns at least gold's reward, with a mean baseline the lowest-probability wrong action gets the largest advantage, and at two options the signal is zero; in one run it cut accuracy on ordered-scale questions from .636 to .111. Scoring the probability as a loss by the pathwise gradient avoids the collapse, as do the REINFORCE rewards r and r minus p. Second, with full labels, none of four tested bandit arms beat cross-entropy plus one temperature on mean expected calibration error (ECE) over four unseen label schemas. Third, from logged feedback alone, Chosen-Action Proper Scoring (CAPS), the direct method with a proper score on the chosen action's probability, beats fine-tuning on confirmed answers only, which discards negatives: on Banking77 by 2.6 accuracy points, .053 lower ECE and lower selective risk (three seeds against one baseline seed), and by 21.7 points on one seed on emotion data with an often-wrong logger. At N 5000, CAPS stays 1.5 points below full-label training at 0.6B and .6 points at 1.7B against one full-label seed, and propensity weights add nothing on this log. Under feedback flipped at a known rate, one seed per rate, CAPS's confidence tracks the noisy channel on average, backward correction recovers most of the clean calibration, and early stopping on held-out logged outcomes prevented memorising noisy draws in all three runs. Code, adapters, logs and per-record results are public.
\end{abstract}


\section{Introduction}\label{sec:1}

\textbf{The decision head.} A decision head answers typed questions about a text state: yes-or-no questions, choice questions among $K$ options, and score questions, questions whose answer is a level on an ordered scale. The options are part of each request, so the same model answers label sets it was never trained on. It returns a full distribution per question from the option-letter probabilities at an answer slot, in one forward pass, with no text generated. The caller's code turns the probabilities into actions, for example by answering above a confidence threshold and escalating below it. That is selective classification with a reject option, and it only works if the probabilities are calibrated and rank the errors lowest. Ours is a LoRA adapter on Qwen3-0.6B-Base and 1.7B-Base, trained by supervised fine-tuning (SFT). Its in-domain expected calibration error (ECE, \eqref{eq:ece} in Appendix~\ref{sec:setup}) is .013 to .015, but it is overconfident on unfamiliar schemas, with ECE up to .22 on the unseen-schema test splits. The setting is the one TypeSafe AI announced with Jev \citep{typesafe2026jev}. Jev is trained by an unpublished recipe called RLCD, reinforcement learning for calibrated decisions. Our decision head is an independent re-implementation of the concept; jevmark is the name of its code repository and released adapters.

\textbf{Why logged bandit feedback is the realistic case.} After deployment on a new domain, labels are expensive and rare. What a system does record is whether the action it took turned out right: a ticket reopened or not, a routing corrected or not. This is logged bandit feedback: one outcome per interaction, only for the chosen action, under a known logging policy. The common practice is to fine-tune on the interactions that went well and discard the rest. That throws the negatives away and, as we show, leaves the model overconfident (Section~\ref{sec:4.3}).

\textbf{This paper.} It makes three claims, the third of which is the core.
\begin{enumerate}
  \item \textbf{A corollary on detached proper-score rewards.} The general principle is that a detached, parameter-dependent reward drops the pathwise term of the gradient \citep{schulman2015scg,mohamed2020montecarlo}. Our contribution is the specific algebra when the reward is the Brier or the log score, in the standard form used here, of the sampled action's own probability. A wrong action's reward is never below gold's, so the lowest-probability wrong action sampled gets the largest advantage. The expected gradient on the gold logit is never positive, and the signal is exactly zero at $K = 2$ (Section~\ref{sec:3}). The contribution also includes the measured collapse (Section~\ref{sec:4.2}). The result is distinct from proper-score rewards on a separately emitted probability, which work \citep{bereket2025uncalibrated,damani2025rlcr}, and from GRPO's standard-deviation normalisation, which our baselines do not use. It is also distinct from OpenJev-RLCD's gradient-norm argument \citep{gao2026openjev}, which concerns a sampled rationale (Section~\ref{sec:2}).
  \item \textbf{A result consistent with a concurrent ablation: with full labels, none of the four bandit arms tested beats cross-entropy plus one temperature on mean ECE over the four unseen label schemas.} Our full-label results are consistent with OpenJev-RLCD's direct-scorer ablation (Section~\ref{sec:2}), which is the one point of agreement. They neither confirm nor contradict OpenJev-RLCD's headline, that two-stage RLCD with rationales beats SFT plus temperature in accuracy and in the area under the risk-coverage curve (AURC). We add the transfer-to-unseen-schema protocol, the continued cross-entropy control, the four bandit arms (REINFORCE with reward $r$ and with reward $r - p$, CAPS-Brier and CAPS-log) under one temperature, and accuracy on the same splits. We also find that a single temperature is the strongest cheap fix. Given its own temperature, the best arm was level with SFT plus one temperature in mean ECE over the four unseen schemas, $-$.002 [$-$.006, +.004] (Section~\ref{sec:4.2}). The direction is not new.
  \item \textbf{From logged feedback alone, CAPS beats positive-only fine-tuning from 2000 interactions, against one positive-only training seed.} CAPS is the direct method of logged bandit feedback \citep{dudik2011doubly}, trained with a proper score on the chosen action (Section~\ref{sec:2}). With the log score it is positive-only fine-tuning plus an unlikelihood term on every refuted option. What is new is the package for LoRA-adapted LLM decision heads at 2000 interactions:
  \begin{itemize}
    \item positive-only fine-tuning, beaten from 2000 interactions on Banking77 and from 500 on emotion, with better calibration and AURC at every log size. Positive-only fine-tuning selects towards the starting model's own successes and loses the exclusions the negatives carry. At N 5000 on Banking77, CAPS-Brier leads it by 2.6 points [1.9, 3.3] in accuracy, a three-seed mean against one positive-only SFT seed (per seed, 2.4 to 3.0). It stays 1.5 points [1.0, 2.0] below full-label training at 0.6B, between three-seed means. At 1.7B, against one full-label seed, it is level on seed 0, $-$.4 points [$-$1.0, +.2], and .6 points [.2, 1.1] below as a seed mean;
    \item for the caller who thresholds, fewer confident errors: at a confidence threshold of .95 on Banking77 at N 5000, seed 0, positive-only fine-tuning keeps .916 of answers at .942 accuracy, so it passes on about 6 percent errors as confident, and CAPS-Brier keeps .743 at .986, with no stored interval (Section~\ref{sec:4.3});
    \item on this log, no gain from propensity weights, clipped or unclipped, the expected behaviour of a well-specified regression \citep{shimodaira2000covariate};
    \item without correction, a chosen-action probability that tracks the noisy channel's rate on average, and backward correction \citep{natarajan2013noisy,patrini2017losscorrection} with its $1/(1-2f)$ factor \citep{angluin1988noisy};
    \item memorisation of a fixed noisy log, and checkpoint selection on the held-out log-likelihood of logged outcomes, which needs no label.
  \end{itemize}
\end{enumerate}

\textbf{What the paper does not show.} The feedback is simulated from gold labels, cleanly or with symmetric flips at a known rate. There are two adaptation domains, Banking77 and emotion, with one logging policy per domain, the starting model itself, and a second log draw on Banking77 as a robustness check. Positive-only SFT has one training seed everywhere and full-label SFT one on emotion and at 1.7B, so those differences are seed means against one baseline seed. The emotion lead of 21.7 points is on one seed, and the noise sweep has one seed per rate. Every interval is a record bootstrap with no training-seed variance. There is no counterfactual-risk policy learner (POEM or BanditNet), no doubly robust arm and no off-policy value estimate of the learned policy (Section~\ref{sec:5}).

\section{Related work}\label{sec:2}

\textbf{Calibrated decision heads for agents.} Jev \citep{typesafe2026jev} answers typed questions about a state with calibrated probabilities. Another published reconstruction, OpenJev-RLCD \citep[arXiv:2609.38850]{gao2026openjev}, reads the softmax over option tokens after an answer marker. OpenJev-RLCD's first public version is arXiv v1, 30 September 2026. In this project's public repository, jevmark, the v1 and v2 results were committed on 26 and 28 September. Their git tags v1 and v2 were created only on 3 October, together with v3. The logged-feedback experiments were committed in the first days of October (tag v3, 3 October). The later experiments were written after reading OpenJev-RLCD: the off-policy baselines, emotion, 1.7B and the noise sweep (tag v4-paper, 17 October). So was the positioning of claims 1 and 2. These commit and tag dates are self-reported from the public repository's history. A head of this kind serves a cascade, in which a learned scorer decides whether a cheaper model's answer is acceptable \citep{chen2023frugalgpt}. Confidence tokens fine-tuned into a small model can drive routing and rejection \citep{chuang2025selfref}. Thresholding the top posterior is selective classification \citep{geifman2017selective}, so we evaluate the caller by its risk-coverage curve. Option-letter readout is biased towards specific option letters \citep{zheng2024selectors}, so we randomise option order in every split. After supervised training, accuracy by gold letter position on the in-domain test split ranges from .939 to .986 over the 14 positions at 0.6B (project report R1, Section~3; Appendix~\ref{sec:B}). Unlike a confidence stated in words \citep{xiong2024uncertainty}, ours is the distribution over the options the head read, with no text generated. JevAdvBench \citep{hu2026jevadvbench} is an adversarial benchmark for typed decision models; it is not used here.

\textbf{Learning from logged bandit feedback.} CAPS is the direct method of logged bandit feedback, which fits a reward model to the logged actions' outcomes and acts greedily on it. With a square loss it appears in \citet{beygelzimer2009offset} and \citet{dudik2011doubly}, and it is the regression-oracle objective \citep{foster2018regcb,foster2020squarecb}. Its known weakness is bias when the reward model is misspecified, which doubly robust estimators correct \citep{dudik2011doubly}. This is why the well-specified case carries weight in Section~\ref{sec:3}. A contextual-bandit bake-off on more than 500 supervised datasets converted to bandit feedback found a greedy regression-oracle learner competitive \citep{bietti2021bakeoff}. That learner fits the observed rewards without importance weights. \citet{strehl2010logged} study learning from the logs of a non-uniform production policy. Counterfactual risk minimisation instead weights the policy objective by inverse propensities, with variance penalties and self-normalisation \citep{swaminathan2015crm,swaminathan2015selfnormalized,joachims2018banditnet}. Simulating bandit feedback from a labelled multiclass dataset is the standard protocol \citep{beygelzimer2009offset}. Importance weights are unnecessary for a well-specified conditional model \citep{shimodaira2000covariate}. Their effect on deep networks vanishes as training proceeds to convergence \citep{byrd2019importance}. Selection stops our clean learners at 1.4 to 3.9 epochs at N 5000 (Table~\ref{tab:A1c}), short of convergence, so our no-gain result rests on its own measurements. Fine-tuning on confirmed answers only is the decision-head form of self-training on filtered successes \citep{zelikman2022star}. The counterpart for the negatives is unlikelihood training, which pushes down the probability of refuted outputs \citep{welleck2020unlikelihood}. On the language-model side, KTO \citep{ethayarajh2024kto} aligns a generator from unpaired binary feedback, desirable or undesirable, with a prospect-theoretic utility against a reference model, and it uses the negatives, so positive-only fine-tuning is not the only practice. KTO scores whole generations against a reference; CAPS is a proper-score regression on one component of a readout distribution, and CAPS-log is the decision-head form of using both signals, cross-entropy on confirmed answers and unlikelihood on refuted ones. We did not run KTO; a decision-head variant is future work. This paper takes the direct method to LoRA-adapted LLM decision heads at 500 to 5000 interactions and compares it with positive-only fine-tuning, the practice it would replace.

\textbf{Proper scoring rules as losses and as rewards.} A strictly proper score is minimised in expectation by the true probability \citep{brier1950verification,gneiting2007proper}. That makes it a calibration-respecting loss and, when the forecast is a separately emitted probability, a sound reward. Throughout, a score is a loss to minimise, and a reward built from it is its negative plus a constant. For an outcome $r \in \{0, 1\}$ and a forecast $p$, the two rewards are the Brier reward $1 - (r - p)^2$ and the log reward $r \log p + (1 - r) \log(1 - p)$. Binned ECE \citep{naeini2015ece} is biased, which is why we report Brier and NLL beside it \citep{nixon2019measuring}, and one temperature repairs calibration post hoc \citep{guo2017calibration}. As RL rewards on a verbalized confidence, proper scores yield calibrated models \citep{damani2025rlcr}. \citet{bereket2025uncalibrated} show that the estimator, not the score, then decides calibration. An unbiased group baseline gives calibrated forecasts, and GRPO's standard-deviation normalisation drives them to the extremes.

\textbf{Calibration in RL for language models and the gradient question.} That a detached, parameter-dependent reward drops the pathwise term of the gradient is the general rule of stochastic computation graphs \citep{schulman2015scg,mohamed2020montecarlo}. Our first result is its corollary for the Brier and log scores of the sampled action's own probability under REINFORCE \citep{williams1992reinforce} with the group baseline of GRPO \citep{shao2024deepseekmath}. Our group baseline subtracts the group mean only, with no standard-deviation normalisation in any full-label run (Appendix~\ref{sec:B}), and the logged-feedback REINFORCE arm subtracts a batch mean only. The failure mode of \citet{bereket2025uncalibrated} therefore does not apply, and our pathology is specific to scoring the sampling probability itself. Outcome-only rewards degrade calibration \citep{damani2025rlcr}. OpenJev-RLCD \citep{gao2026openjev} optimises the readout distribution by the pathwise gradient of a proper score, its deterministic policy gradient with respect to the reported distribution. It finds the score-function term, which acts on a sampled rationale, at 7 to 12 times the pathwise term's gradient norm, and more of it monotonically worse. It thus reaches the same recipe by a gradient-norm argument, but about a different object: its REINFORCE term acts on a sampled rationale, not on the answer whose probability is scored. Its direct-scorer ablation reports that proper-score RL never beat cross-entropy on CLINC150 (94.6 against 96.3) and that RLVR collapsed to the mode. Our second result is consistent with that ablation, under a transfer protocol with a continued cross-entropy control.

\textbf{Noisy feedback.} Feedback flipped at a known rate $f$ is random classification noise on the outcome \citep{angluin1988noisy}. The observed outcome rate that CAPS fits, $f + (1-2f)\,P(\mathrm{correct})$ in \eqref{eq:channel}, is the symmetric case of the class-conditional identity \citep{natarajan2013noisy,menon2015corrupted}. In that identity the clean posterior is an affine transform of the corrupted one. Inverting it is backward correction with a known transition matrix \citep{patrini2017losscorrection}, whose error grows by the $1/(1-2f)$ factor of \citeauthor{angluin1988noisy}. \citet{agarwal2021noisybandit} study flipped bandit feedback with a known rate for an online linear learner. Networks memorise noisy labels given enough passes \citep{zhang2017rethinking}, and early stopping is the provable remedy \citep{li2020earlystopping}.

\textbf{Calibration under shift and after fine-tuning.} Calibration of a language model's own token probabilities on multiple-choice questions has been studied for large pretrained models \citep{kadavath2022know} and for question answering \citep{jiang2021know}. The bias of option probabilities can be corrected post hoc from content-free inputs \citep{zhao2021calibrate}, and \citet{kamath2020selective} train a selective question answerer to abstain under domain shift, the setting of our unseen-schema splits. A single temperature is strong in-domain and fragile under shift \citep{ovadia2019shift}. Our second result says the bandit alternatives are no less fragile on unseen label schemas. LoRA forgets less than full fine-tuning but still forgets \citep{biderman2024lora}, so we report forgetting on the original domain.

\section{Method: Chosen-Action Proper Scoring}\label{sec:3}

\textbf{Setting.} A logged interaction is $(x, O, a, q(a \mid x), r)$. Here $x$ is a state, $O$ its $K$ options and $a$ the action the logging policy $q$ chose. The logging propensity $q(a \mid x)$ is recorded exactly, and $r \in \{0, 1\}$ records whether $a$ was right.

The learner starts from a general decision head $p_\theta$ and sees only the log. We write $p_a$ for $p_\theta(a \mid x)$, the option-letter softmax probability of option $a$. In our experiments the logging policy is epsilon-mixture sampling from the starting model's distribution $p_0$, with $\epsilon = 0.1$:
\begin{equation}
  \label{eq:logger}
  q(a \mid x) = 0.9\, p_0(a \mid x) + 0.1/K.
\end{equation}
Every option therefore has support, and the logger is the learner's own starting point.

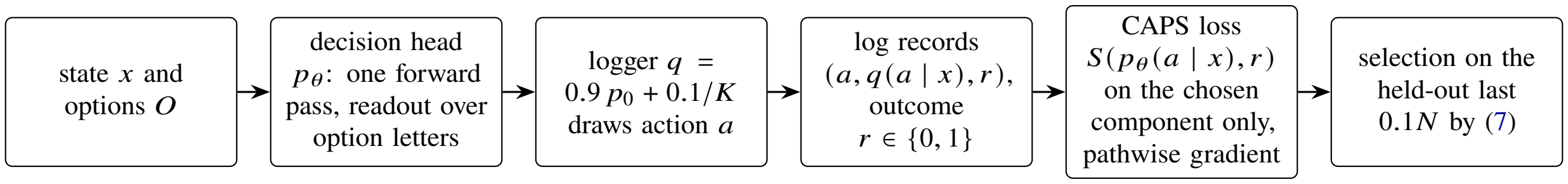
\begin{figure}[t]
\centering
\begin{tikzpicture}[
  box/.style={draw, rounded corners=2pt, align=center, text width=2.2cm, minimum height=1.55cm, inner sep=3pt, font=\footnotesize},
  arr/.style={-{Stealth[length=2mm]}, semithick},
  node distance=0.34cm]
\node[box] (x) {state $x$ and options $O$};
\node[box, right=of x] (head) {decision head $p_\theta$: one forward pass, readout over option letters};
\node[box, right=of head] (log) {logger $q = 0.9\,p_0 + 0.1/K$ draws action $a$};
\node[box, right=of log] (r) {log records \mbox{$(a, q(a \mid x), r)$}, outcome \mbox{$r \in \{0, 1\}$}};
\node[box, right=of r] (caps) {CAPS loss $S(p_\theta(a \mid x), r)$ on the chosen component only, pathwise gradient};
\node[box, right=of caps] (sel) {selection on the held-out last $0.1N$ by \eqref{eq:select}};
\draw[arr] (x) -- (head);
\draw[arr] (head) -- (log);
\draw[arr] (log) -- (r);
\draw[arr] (r) -- (caps);
\draw[arr] (caps) -- (sel);
\end{tikzpicture}
\caption{The setting. The head returns a distribution over the options in one forward pass. The logger draws one action and records only its outcome and propensity. CAPS fits a proper score to the chosen action's probability against that outcome and selects a checkpoint on held-out logged outcomes; no gold label is read. Provenance: Table~\ref{tab:prov}, F\ref{fig:setting}.}\label{fig:setting}
\end{figure}

\textbf{Objective.} CAPS minimises
\begin{equation}
  \label{eq:caps}
  L(\theta) = \mathbb{E}_x\, \mathbb{E}_{a \sim q(\cdot \mid x)} \bigl[ S\bigl(p_\theta(a \mid x), r(x, a)\bigr) \bigr]
\end{equation}
with $S$ a strictly proper binary score and $r(x, a)$ the outcome of action $a$. The default is Brier, $S(p, r) = (r - p)^2$; the alternative is the log score, $S(p, r) = -r \log p - (1 - r) \log(1 - p)$. We write CAPS-Brier and CAPS-log for the two. The empirical mean over the log estimates $L$ directly. The gradient is the pathwise gradient through $p_a$:
\begin{align}
  \nabla_\theta (r - p_a)^2 &= -2\,(r - p_a)\, \nabla_\theta p_a, \label{eq:grad-brier}\\
  \nabla_\theta \bigl[-r \log p_a - (1 - r) \log(1 - p_a)\bigr] &= -\Bigl(\frac{r}{p_a} - \frac{1 - r}{1 - p_a}\Bigr) \nabla_\theta p_a. \label{eq:grad-log}
\end{align}
A right action is pushed up and a wrong one down. With the log score the loss on a positive is $-\log p_a$, the cross-entropy that positive-only fine-tuning applies to the confirmed answer. On a negative it is $-\log(1 - p_a)$, an unlikelihood term on the refuted option \citep{welleck2020unlikelihood}. CAPS-log is thus positive-only fine-tuning plus the negatives, and CAPS-Brier is its squared-error analogue.

\textbf{Identity.} CAPS is the direct method of \citet{dudik2011doubly}, with a proper score as the regression loss on the chosen action's outcome and the argmax of the fitted head as the policy. It is also OpenJev-RLCD's stage 1 specialised to partial feedback. Stage 1 scores the whole readout distribution against the one-hot label; CAPS scores the one component the log reveals, the chosen action's probability against its own outcome.

\textbf{Why no propensity weight.} For each $(x, a)$ with $q(a \mid x) > 0$, the inner term of \eqref{eq:caps} is minimised at $p(a \mid x) = P(r = 1 \mid x, a)$, whatever $q$ is. The target does not depend on the logging policy as long as it has support, so the unweighted fit is consistent for a well-specified model \citep{shimodaira2000covariate}. Under clean feedback the target $1[a = g]$ over the $K$ options is one-hot, which the softmax represents. Under symmetric flips at rate $f$ the $K$ per-action targets $f + (1 - 2f)\,1[a = g]$ sum to $1 + (K - 2)f$, so for $K > 2$ no softmax matches all of them and the direct method is misspecified; Section~\ref{sec:4.3} bounds what that costs. The propensity only sets how a finite model trades errors between actions. $L$ makes that trade as the logging policy acts, which here is close to the starting model. A weight $1/q$ would retarget it to a uniform average over options. Writing $r_a$ for the outcome action $a$ would receive,
\begin{equation}
  \label{eq:ipsbrier}
  \mathbb{E}_{a \sim q(\cdot \mid x)} \Bigl[ \frac{(r_a - p_a)^2}{q(a \mid x)} \Bigr] = \sum_a (r_a - p_a)^2,
\end{equation}
the multiclass Brier score of the whole readout against the one-hot gold. The logged REINFORCE arm with reward $r - p$ uses the reward $r - p_a$, detached, a micro-batch mean baseline and the weight $\min(p/q, 5)$ (Appendix~\ref{sec:setup}). Per question, and exactly while no weight is clipped, its expected gradient is
\begin{equation}
  \label{eq:reinforce-ips}
  \tfrac{1}{2}\,(1 - 1/B)\, \nabla_\theta \sum_a (r_a - p_a)^2,
\end{equation}
with $B = 8$ the micro-batch size. It is therefore the IPS-weighted CAPS-Brier estimator. A check on a tiny model matches the identity to a relative difference of $4.5 \times 10^{-7}$. Its clip binds for under 1 percent of weights (Section~\ref{sec:4.3}).

Accuracy is a consequence, not the objective: moving right actions up and wrong ones down moves the argmax, and options the logger never takes are unconstrained. A negative on an option other than the logger's top choice arises only from the logger's randomness. On our main log, 1514 of 2593 negatives (.584) fall on such options, and the uniform exploration term's posterior responsibility for them averages .545.

\textbf{Selection.} Train on the first $0.9N$ interactions and hold out the last $0.1N$, the set $V$. From a 50-step grid that excludes step 0 (the starting adapter), keep the checkpoint with the highest mean Bernoulli log-likelihood of the held-out logged outcomes under $p(\text{logged action})$:
\begin{equation}
  \label{eq:select}
  \ell_V(\theta) = \frac{1}{|V|} \sum_{i \in V} \Bigl[ r_i \log p_\theta(a_i \mid x_i) + (1 - r_i) \log\bigl(1 - p_\theta(a_i \mid x_i)\bigr) \Bigr],
\end{equation}
with $(x_i, a_i, r_i)$ the $i$-th logged interaction. This is early stopping on the held-out log score of logged outcomes, and no gold label is read. The rule is the same for every trained learner, positive-only SFT and full-label SFT included. It scores the held-out negatives too, so positive-only SFT's checkpoint is chosen with negative information. Under noisy feedback this is what prevents memorisation (Section~\ref{sec:4.3}).

\begin{corollary}[The Brier or log score of the sampled action's probability as a detached reward]\label{cor:detached}
Take $K \ge 2$ options with one gold option $g$ and probabilities $p_a$ that sum to one over the $K$ options (the option-letter softmax). Use as the reward of the sampled action the Brier or the log score in the standard form used here, with $r_a = 1[a = g]$: Brier, $R(a) = 1 - (r_a - p_a)^2$, or the log score, $R(a) = r_a \log p_a + (1 - r_a) \log(1 - p_a)$.

The gradient of the expected reward, $\sum_a p_a R(a)$, has a score-function term, $\sum_a R(a)\, \nabla p_a$, and a pathwise term, $\sum_a p_a \nabla R(a)$. Detaching $R$ keeps only the first \citep{schulman2015scg}. Form REINFORCE advantages against the mean of a group of $G$ actions sampled independently for the same $x$, either from the current $p$ or from a distribution $q$, held fixed during the step, with $q_a > 0$ for every $a$. Under $q$, each sample's advantage is multiplied by its unclipped weight $p_a / q_a$. The group mean is the unweighted mean of the $G$ rewards and includes the sample itself, and the advantage is not divided by the group standard deviation.

(i) Every wrong action's reward is at least gold's, $R(w) \ge R(g)$. Equality holds at $K = 2$, and the inequality is strict whenever $p_w < 1 - p_g$. In each group, the lowest-probability wrong action sampled has the largest reward, so its advantage is non-negative. Whenever the group also samples gold, gold's advantage is non-positive.

(ii) If $K = 2$, all rewards in every group are equal, and the gradient is zero.

(iii) For every $K$, the expected detached gradient on the gold logit $z_g$, per sample, is
\begin{equation}
  \label{eq:gold}
  (1 - 1/G)\, p_g \Bigl( R(g) - \sum_a p_a R(a) \Bigr) \le 0,
\end{equation}
whether the actions are sampled from $p$ or from $q$ with unclipped weights $p/q$.
\end{corollary}

\begin{proof}
A wrong action $w$ has $r_w = 0$ and $p_w \le 1 - p_g$. Under Brier, $R(w) = 1 - p_w^2$ and $R(g) = 1 - (1 - p_g)^2$; under the log score, $R(w) = \log(1 - p_w)$ and $R(g) = \log p_g$. Both rewards decrease in $p_w$ and equal $R(g)$ at $p_w = 1 - p_g$. Hence $R(w) \ge R(g)$, strictly when $p_w < 1 - p_g$, and the smallest sampled $p_w$ gives the group maximum. At $K = 2$, $p_w = 1 - p_g$ exactly, so $R(w) = R(g)$. For (iii), let $a_1, \dots, a_G$ be drawn independently from $q$ ($q = p$ on policy), with weights $w_k = p_{a_k} / q_{a_k}$ and group mean $\bar R = \frac{1}{G} \sum_j R(a_j)$. The update direction is $\frac{1}{G} \sum_k w_k \bigl(R(a_k) - \bar R\bigr) \nabla \log p_{a_k}$, with $R$ and $w$ detached. For any $h$ that does not depend on $\theta$, $\mathbb{E}\bigl[w_k\, h(a_k) \nabla \log p_{a_k}\bigr] = \sum_a q_a \frac{p_a}{q_a} h(a) \nabla \log p_a = \sum_a h(a) \nabla p_a$, and for a constant $h$ this is $\nabla \sum_a p_a = 0$. Split $\bar R$ into the sample's own term $R(a_k)/G$ and the other $G - 1$ terms. The others are independent of $a_k$, so their expectation multiplies $\mathbb{E}[w_k \nabla \log p_{a_k}] = 0$ and drops out. What remains is $(1 - 1/G) \sum_a R(a) \nabla p_a$. Since $\partial p_a / \partial z_g = p_a (1[a = g] - p_g)$, its component on $z_g$ is $(1 - 1/G)\, p_g \bigl(R(g) - \sum_a p_a R(a)\bigr)$. The group mean itself converges to $\sum_a q_a R(a)$, the $q$-average of the rewards, as $G$ grows. That limit does not enter the expectation: only the sample's own share $1/G$ of the mean does, and it gives the factor $1 - 1/G$ under $p$ and $q$ alike. By (i) every $R(a) \ge R(g)$, so $\sum_a p_a R(a) \ge R(g)$.
\end{proof}

The score grades $p_a$ as a forecast of $r_a$, and a low probability on a wrong option is always a good forecast. CAPS uses the same score as a loss on $p_a$ by its pathwise gradient \eqref{eq:grad-brier}, not as a reward on $a$. The corollary does not contradict proper-score rewards on a confidence the model emits separately \citep{damani2025rlcr,bereket2025uncalibrated}. It applies only to the probability with which the model sampled the rewarded action. It is stated for these two rewards, and propriety alone does not give it: adding a constant that depends on the outcome keeps a score proper, but can reverse the ordering in (i) and make the rewards in (ii) unequal. In the full-label runs $p/q \le 1/0.9$, so the clip of 5 never binds and (iii) holds exactly for their sampler (Section~\ref{sec:4.2}).

\section{Experiments}\label{sec:4}

\subsection{Setup}\label{sec:4.1}

\textbf{Model and data.} The decision head is Qwen3-0.6B-Base or 1.7B-Base with LoRA \citep{hu2022lora,qwen2025qwen3}, read from the letter logits at each question's answer slot. It is trained with supervision on CLINC150 intents \citep{larson2019clinc} and SST-5 sentiment \citep{socher2013sst}, giving the general decision heads, the 0.6B and 1.7B SFT heads. It is evaluated on the in-domain test split (the CLINC150 test questions), the SST-5 test split, the unseen-intents test split (20 held-out CLINC150 intents) and four schemas never trained on. These are AG News, Yelp \citep{zhang2015charcnn}, emotion \citep{saravia2018carer} and Banking77 \citep{casanueva2020banking77}. Score questions come from SST-5 and Yelp only, and the in-domain test split has none. SST-5 sentiment is on a five-level scale in 70 percent of records and a three-level one in 30 percent; Yelp stars, an unseen score schema, are on a five-level scale. Training details and the data gates are in Appendix~\ref{sec:setup}.

\textbf{Adaptation protocol.} The 0.6B SFT head logs every message of the Banking77 train split once, 9942 messages, in a fixed seeded shuffle of the file. The file is grouped by intent, and the shuffle makes every prefix and every selection set a random sample. Each message is a 10-option choice: gold, 8 distractors drawn uniformly from the other 76 intents and \texttt{other}, in a seeded order per record. The test split is built the same way, so every Banking77 number is a 10-way accuracy, not comparable with published 77-way results. Gold is never \texttt{other}, so \texttt{other} is always wrong. The logger chose it in .070 of interactions, 698 of the 2593 negatives (26.9 percent). Of those, 371 came when \texttt{other} was the logger's top choice and count among the negatives on the top option, and 327 count among the 1514 off it (Section~\ref{sec:3}). A rule against \texttt{other} is therefore learnable from the negatives alone. Positive-only SFT, which never sees \texttt{other} confirmed, learns it only indirectly; emotion has no \texttt{other} option. The logger samples from $q$ in \eqref{eq:logger} and records only the correctness of its action. Its top choice is right .836 of the time (its policy accuracy), and its sampled action .739 of the time (the outcome rate). The logger is the starting model itself, so positive-only SFT trains on the starting model's own successes. The 0.6B SFT head is thus at once the starting model, the logger and the zero-shot reference. A logger other than the starting model would change what the positives contain. At 1.7B, the 1.7B SFT head draws its own log of the same messages, with policy accuracy .837 and outcome rate .753. Learners see the first $N \in \{500, 2000, 5000\}$ interactions. Every trained learner takes the same number of steps, $\max(200, \lceil 10 \times 0.9 N / 32 \rceil)$, so a learner that keeps only the positives makes more passes over them. That is 13.4 epochs for positive-only SFT at N 5000 against 10.0 for the others (selected steps in Table~\ref{tab:A1c}). On Banking77 at 0.6B and N 5000, CAPS-Brier, CAPS-log and full-label SFT each have three training seeds, and CAPS-Brier also has a second log draw. At 1.7B and on emotion, CAPS-Brier has three seeds at N 5000 and full-label SFT one. Positive-only SFT has one seed everywhere, and every learner has one at N 500 and 2000. Evaluation is on the full Banking77 test split (3080 records), plus the in-domain and unseen-intents test splits for forgetting. Intervals are paired 95 percent record-bootstrap intervals \citep{efron1979bootstrap,efron1993introduction}. These resample test records and carry no training-seed variance. Beside every seed-mean difference we therefore give the seed range, and where the ranges overlap the difference is unresolved at three seeds.

\textbf{Learners.} The temperature learner fits one temperature to the logged outcomes and changes no argmax. Positive-only SFT is cross-entropy on confirmed actions, a common practice. It selects towards the interactions the starting model already gets right and loses the exclusions the negatives carry. In clean simulation a positive reveals the gold label (Section~\ref{sec:5}). Full-label SFT is cross-entropy on gold for every logged interaction, the upper bound. Three off-policy forms use the propensities: the logged REINFORCE arm with reward $r - p$, IPS CAPS-log and IPS positive-only SFT (Section~\ref{sec:3} and Appendix~\ref{sec:setup}). They are not counterfactual risk minimisation, which weights the policy objective itself and was not run (Section~\ref{sec:5}). CAPS-Brier and CAPS-log name the CAPS learners throughout. Run names, file names and split names appear only in Appendix~\ref{sec:B}.

\textbf{Metrics.} We report accuracy, ECE with 15 equal-width bins, and the multiclass Brier score and negative log-likelihood (NLL) of the whole readout against the gold option (Appendix~\ref{sec:setup}). Two selective-classification metrics serve the caller who thresholds \citep{geifman2017selective}, with questions ranked by top-1 probability. AURC is the area under the risk-coverage curve \eqref{eq:aurc}. Cov@eps is the largest share of questions a top-1 threshold keeps with error at most eps \eqref{eq:cov}. Lower AURC and higher Cov@eps are better, and the tables show Cov@.05; the definitions are in Appendix~\ref{sec:setup}. Binned ECE \eqref{eq:ece} is biased upwards under resampling, so its bootstrap intervals sit mostly above the point estimate. An example is .023 [.021, .036] for the temperature learner at N 500.

A temperature changes the level of the probabilities and only slightly their ranking. At N 5000 the temperature learner cuts zero-shot ECE from .063 to .020 but moves AURC only from .0333 to .0328 (Table~\ref{tab:A1}; details in Appendix~\ref{sec:A1}).

\subsection{Full labels: the negative results}\label{sec:4.2}

A first run used Brier as a detached REINFORCE reward on fully labelled in-domain data. It used the sampler of the two REINFORCE arms below: $G = 4$ actions per question from $q = 0.9\,p + 0.1/K$, each weighted by $p/q$ clipped at 5. The baseline was the group mean without standard-deviation normalisation. The loss added $\beta\,\mathrm{KL}(p \,\|\, p_{\mathrm{SFT}})$ with $\beta = .02$, where $p_{\mathrm{SFT}}$ is the 0.6B SFT head. On score questions, accuracy fell from .636 to .111 on SST-5 and from .505 to .044 on Yelp, one run. Accuracy on the in-domain test split, CLINC150 only, with no score questions, fell only from .956 to .948 (Table~\ref{tab:A6b}). A simulation of the same sampler, weights and baseline ran on the 0.6B SFT head's training questions, with no KL term, whose gradient is zero at the starting model. It lowered the gold logit in 24.0 percent of questions, raised it in 1.8 percent and left it unchanged in 74.2 percent. Almost all unchanged questions are groups of four identical samples (63.7 percent) or of distinct samples with equal rewards (10.4 percent). The latter include every yes-or-no question, as $K = 2$ implies. Since the expected gradient on the gold logit is never positive (Corollary~\ref{cor:detached}(iii), \eqref{eq:gold}), the 1.8 percent are finite-group noise (Table~\ref{tab:A6c}).

Five arms then ran three seeds each at 0.6B on full labels. Each run takes 500 steps with warmup and is evaluated at the checkpoint, from a grid every 100 steps, with the lowest NLL on a 1000-record gold-labelled validation subset; the detached-Brier run selected step 100. This selection reads gold labels, unlike the label-free rule \eqref{eq:select} of Section~\ref{sec:3}, so every arm here, like the temperature of SFT, is chosen with gold (Appendix~\ref{sec:setup}). Each bandit arm samples $G = 4$ actions per question from $q = 0.9\,p + 0.1/K$ and sees $r_a = 1[a = g]$ for the sampled actions only. The policy-gradient arms weight by $p/q$, which never reaches its clip of 5 since $p/q \le 1/0.9$. Every arm but continued cross-entropy adds $\beta\,\mathrm{KL}(p \,\|\, p_{\mathrm{SFT}})$ with $\beta = .02$. In expectation, per question, before the KL term and with $q$ held fixed, the arms follow the gradients of
\begin{equation}
  \label{eq:arms}
  \begin{aligned}
    &\text{continued cross-entropy:} && -\log p_g,\\
    &\text{REINFORCE, outcome reward:} && -(1 - 1/G)\, p_g,\\
    &\text{REINFORCE, reward } r - p\text{:} && \tfrac{1}{2}(1 - 1/G) \textstyle\sum_a (r_a - p_a)^2,\\
    &\text{CAPS-Brier:} && \textstyle\sum_a q_a (r_a - p_a)^2,\\
    &\text{CAPS-log:} && \textstyle\sum_a q_a \bigl[-r_a \log p_a - (1 - r_a) \log(1 - p_a)\bigr],
  \end{aligned}
\end{equation}
with $q_a = q(a \mid x)$. The outcome reward raises gold's probability with no calibration term. The reward $r - p$ gives the unweighted multiclass Brier score, and CAPS-Brier the same Brier score weighted by $q$. CAPS-log is cross-entropy on gold plus unlikelihood on the other options, weighted by $q$. REINFORCE with $r - p$ and CAPS-Brier are thus one objective under two weightings.

\textbf{Accuracy.} Every arm recovers almost all of the broken run's loss. The two REINFORCE arms recover as much as the pathwise ones (.623 and .624 on SST-5 score questions against CAPS-Brier's .622), so accuracy does not single out the pathwise gradient. The two REINFORCE rewards tested, $r$ and $r - p$, avoid the collapse. What the pathwise loss keeps is the proper score's calibration meaning, a loss on the probability itself. CAPS-Brier and CAPS-log keep in-domain accuracy (.957 and .958 against .956). They stay .9 to 1.4 points below the SFT head on SST-5 score questions, $-$.014 [$-$.023, $-$.005] and $-$.009 [$-$.017, $-$.001]. These are seed means against the SFT head's one run, with per-seed differences of $-$.016 to $-$.010 and $-$.016 to $-$.003. Continued cross-entropy loses as much, $-$.010 [$-$.019, $-$.002]. On Yelp, CAPS-Brier is level, +.001 [$-$.008, +.012]. CAPS-log's $-$.011 [$-$.021, $-$.001] is unresolved at three seeds (per seed, $-$.024 to +.001) (Table~\ref{tab:A6b}).

\begin{figure}[t]
\centering
\includegraphics[width=\textwidth]{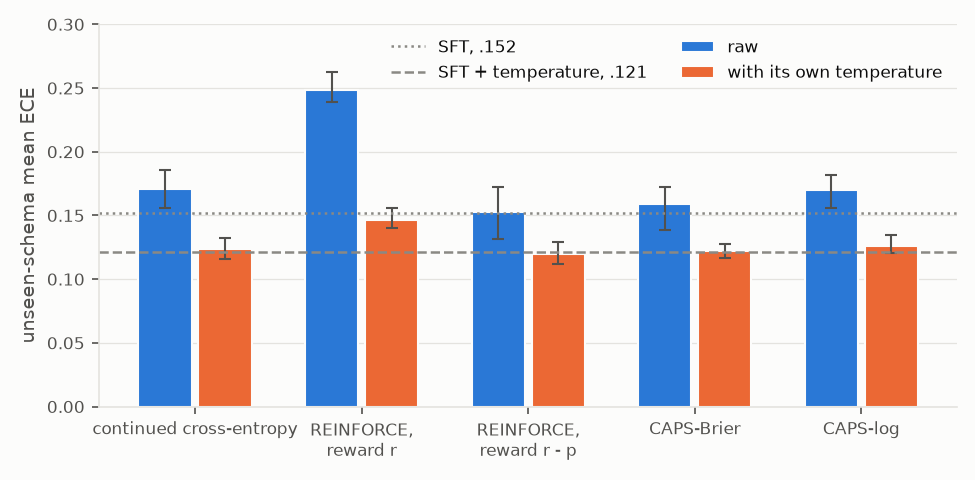}
\caption{Unseen-schema mean ECE per full-label arm, 0.6B: the mean of three seeds with their range, raw and with the arm's own temperature. That temperature is fitted on the in-domain validation split, and the references are SFT and SFT plus one temperature. Provenance: Table~\ref{tab:prov}, F\ref{fig:v2_arm_ece}.}\label{fig:v2_arm_ece}
\end{figure}

\textbf{Calibration.} SFT plus one in-domain temperature has unseen-schema mean ECE .121. Every arm's unseen-schema ECE is higher at every seed, by +.031 to +.127 as seed means against one SFT run (per-seed ranges in Table~\ref{tab:A6}). Given its own temperature, fitted like SFT's on the in-domain validation split, the best arm and the control are only level in mean ECE over the four unseen schemas (Figure~\ref{fig:v2_arm_ece}). When gold is known, whether a sampled answer was right carries no more than the label itself, and one temperature is the strongest cheap fix. This is consistent with OpenJev-RLCD's direct-scorer ablation on CLINC150, which reports accuracy with a different training setup. The transfer to unseen schemas and the continued cross-entropy control are ours.

\textbf{Takeaway.} With full labels, every bandit arm avoids the collapse, and none of the four beats SFT plus one temperature on mean unseen-schema ECE.

\subsection{Logged bandit feedback only: Banking77}\label{sec:4.3}

\begin{table}[t]
\centering
\small
\caption{Summary at N 5000. Three-seed mean (range) where three seeds exist, otherwise seed 0 ($\dagger$). Intervals and paired differences are in Appendix~\ref{sec:A}; every interval there is a record bootstrap with no training-seed variance.}
\label{tab:main}
\begin{tabular}{@{}lrrr@{}}
\toprule
learner & accuracy & ECE & AURC \\
\midrule
\multicolumn{4}{@{}l}{\textit{Banking77, 0.6B}} \\
\quad zero-shot & .851 & .063 & .0333 \\
\quad temperature & .851 & .020 & .0328 \\
\quad positive-only SFT & .905$^{\dagger}$ & .078$^{\dagger}$ & .0202$^{\dagger}$ \\
\quad CAPS-Brier & .931 (.929, .935) & .025 (.019, .031) & .0107 (.0104, .0110) \\
\quad full-label SFT & .946 (.943, .949) & .022 (.016, .030) & .0067 (.0063, .0071) \\
\addlinespace
\multicolumn{4}{@{}l}{\textit{Banking77, 1.7B}} \\
\quad zero-shot & .848 & .089 & .0350 \\
\quad temperature & .848 & .031 & .0333 \\
\quad positive-only SFT & .913$^{\dagger}$ & .067$^{\dagger}$ & .0177$^{\dagger}$ \\
\quad CAPS-Brier & .947 (.942, .949) & .016 (.013, .021) & .0073 (.0066, .0085) \\
\quad full-label SFT & .953$^{\dagger}$ & .023$^{\dagger}$ & .0053$^{\dagger}$ \\
\addlinespace
\multicolumn{4}{@{}l}{\textit{emotion, 0.6B}} \\
\quad zero-shot & .540 & .251 & .295 \\
\quad temperature & .540 & .024 & .294 \\
\quad positive-only SFT & .615$^{\dagger}$ & .256$^{\dagger}$ & .233$^{\dagger}$ \\
\quad CAPS-Brier & .827 (.821, .832) & .046 (.031, .070) & .059 (.058, .059) \\
\quad full-label SFT & .871$^{\dagger}$ & .041$^{\dagger}$ & .031$^{\dagger}$ \\
\bottomrule
\end{tabular}
\par\smallskip{\footnotesize $^{\dagger}$One training seed.\par}
\end{table}

\begin{figure}[t]
\centering
\includegraphics[width=\textwidth]{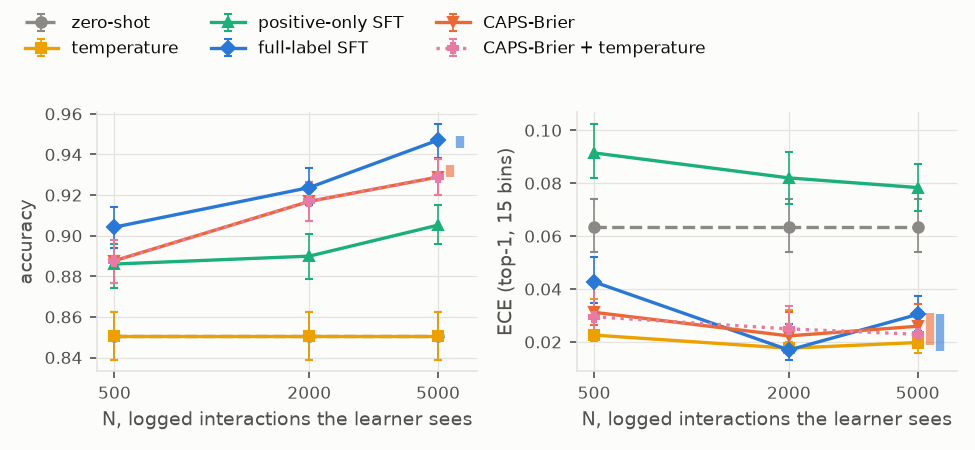}
\caption{Accuracy and ECE against N on the Banking77 test split, 0.6B. Every learner is shown at seed 0 with 95 percent record-bootstrap intervals, and CAPS-Brier also with a temperature fitted on its own training interactions. At N 5000 the range over three training seeds is shown for CAPS-Brier and full-label SFT. Provenance: Table~\ref{tab:prov}, F\ref{fig:v3_n_curve}.}\label{fig:v3_n_curve}
\end{figure}

\textbf{The N-curve.} From 2000 interactions CAPS-Brier beats positive-only SFT (Table~\ref{tab:main} summarises N 5000). It ties at N 500, +.002 [$-$.006, +.009], and leads by +.027 [+.019, +.035] at N 2000 and +.024 [+.016, +.031] at N 5000. Its ECE is lower by .052 to .060 and its AURC by .0035 to .0099 at every N. Cov@.05 is level at N 500 and higher by +.065 [+.048, +.093] and +.069 [+.046, +.094] at N 2000 and 5000 (Figure~\ref{fig:v3_n_curve}; Table~\ref{tab:A1b}). Against full-label SFT it is 1.7 and 1.8 points below in accuracy at N 500 and 5000 and level at N 2000, $-$.007 [$-$.014, +.000]. Its AURC is .0026 to .0047 above full-label SFT's. The temperature learner keeps zero-shot accuracy (.851). It has the lowest ECE at N 500 and 5000 and is within .001 of the lowest at N 2000 (.018 against full-label SFT's .017). Its Brier, NLL and AURC, however, stay far above every trained learner's.

\textbf{Seeds.} Over three CAPS-Brier seeds at N 5000, the seed mean is +.026 [+.019, +.033] above positive-only SFT in accuracy, a three-seed mean against one baseline seed. In AURC it is $-$.0095 [$-$.0121, $-$.0070]. Against full-label SFT's three-seed mean it is $-$.015 [$-$.020, $-$.010] in accuracy, with seed ranges .929 to .935 against .943 to .949. In ECE it is +.003 [$-$.001, +.008], with seed ranges .019 to .031 against .016 to .030, so ECE is unresolved at three seeds (Table~\ref{tab:A2b}). Training seeds and the second log draw move accuracy by at most .6 points. A temperature fitted on CAPS-Brier's own training interactions comes out near 1 and moves ECE by at most .003 (Appendix~\ref{sec:A2}).

\textbf{Takeaway.} On Banking77, against one positive-only SFT seed, CAPS is more accurate from N 2000 and better calibrated at every N, and its gap to full-label SFT does not shrink with N.

\textbf{Off-policy baselines.} On this log and with these estimators, propensity weights add nothing (Figure~\ref{fig:v4_offpolicy}; Table~\ref{tab:A3}). IPS positive-only SFT equals positive-only SFT, +.000 [$-$.003, +.004] in accuracy at N 5000. IPS CAPS-log and IPS positive-only SFT share one weight, $\min(1/q, 5)$ over its mean on the training prefix $P$, the first $0.9N$ interactions:
\begin{equation}
  \label{eq:ipsw}
  w_i = \frac{\min\bigl(1/q(a_i \mid x_i),\, 5\bigr)}{\frac{1}{|P|} \sum_{j \in P} \min\bigl(1/q(a_j \mid x_j),\, 5\bigr)}, \qquad i \in P.
\end{equation}
The clip binds for 12 to 12.5 percent of prefix weights, almost all of them negatives: 43.2 and 46.3 percent of the negatives at N 2000 and 5000 against about 1 percent of the positives, so the clipped arms compress the weights on the negatives (counts in Table~\ref{tab:A3} and Appendix~\ref{sec:setup}). Before normalisation the weights run from 1.1, the floor set by $q \le 0.9 + 0.1/K$, to the cap of 5. The logged REINFORCE arm with reward $r - p$, the IPS-weighted CAPS-Brier estimator of Section~\ref{sec:3}, clips under 1 percent of its weights over all training steps (.7 and .5 percent). It is level with CAPS-Brier, $-$.001 [$-$.006, +.003] at both N. The no-weight finding thus rests on a weighted-Brier arm whose clip binds for under 1 percent of weights, as well as on the clipped IPS arms. No off-policy arm beats CAPS-Brier on accuracy with an interval excluding 0. With full support and a well-specified regression target, this is the expected result. The log score in place of Brier is a choice within the same family. Over three seeds each, CAPS-log minus CAPS-Brier is +.002 [$-$.001, +.005] in accuracy and $-$.020 [$-$.029, $-$.010] in NLL. The seed ranges overlap: .931 to .934 against .929 to .935 in accuracy, and .239 to .275 against .2737 to .2742 in NLL. The difference is therefore unresolved at three seeds, and Brier stays the default (Appendix~\ref{sec:A3}).

\textbf{Takeaway.} On this log, the unweighted direct method is enough: no propensity-weighted arm beats it on accuracy with an interval excluding 0.

\begin{figure}[t]
\centering
\includegraphics[width=\textwidth]{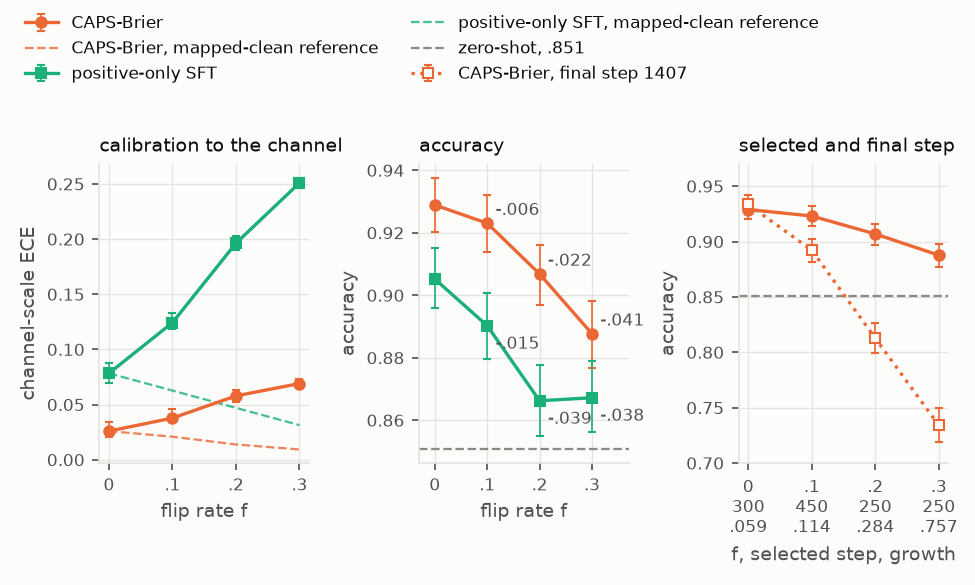}
\caption{Noise-rate sweep on the Banking77 log, 0.6B, N 5000, seed 0, one seed per rate with nested draws. Circles are CAPS-Brier and squares positive-only SFT, and every whisker is a 95 percent record-bootstrap interval. Left: channel-scale ECE, the stored top-1 probability scored against the channel's outcome rate $f + (1-2f) \times \mathrm{correct}$; at f 0 it is the ordinary ECE. The dashed coloured lines are each learner's mapped-clean reference, its clean run's top-1 probability mapped through $f + (1-2f)\,p$ and scored the same way. Centre: accuracy, each noisy point labelled with its paired accuracy loss from the clean run, whose interval is in Table~\ref{tab:A4}; the dashed grey line is zero-shot accuracy, .851. Right: CAPS-Brier's accuracy at the selected step (filled circles, solid line) and at the final step 1407 (open squares, dotted line), with zero-shot dashed. Each tick gives the selected step and the criterion growth. The best achievable value of the selection criterion \eqref{eq:select} is $-H(f)$ for flipped outcomes, with $H$ the binary entropy in nats, and 0 for clean ones. Its deterioration is that value minus the criterion, and criterion growth is the deterioration at the final step minus at the selected step. Provenance: Table~\ref{tab:prov}, F\ref{fig:v4_noise_sweep}.}\label{fig:v4_noise_sweep}
\end{figure}

\textbf{Noisy feedback.} Outcomes are flipped at $f \in \{0.1, 0.2, 0.3\}$ with nested draws. One uniform draw per interaction decides the flip at every rate, so an outcome flipped at 0.1 is flipped at 0.2 and 0.3 too, and the rates are paired. There is one seed per rate, three runs per learner. With the logged outcome $\tilde r$ flipped at rate $f$, the outcome rate of action $a$ is
\begin{equation}
  \label{eq:channel}
  P(\tilde r = 1 \mid x, a) = f + (1 - 2f)\, P(\mathrm{correct} \mid x, a),
\end{equation}
with $P(\mathrm{correct} \mid x, a)$ the clean probability that $a$ is right. Without correction, the chosen action's probability tracks the channel rate on average: the selected run at f 0.2 has mean top-1 probability .743 against a channel line of .744 (Figure~\ref{fig:v4_noise_sweep}; Table~\ref{tab:A4b}). A softmax head cannot match \eqref{eq:channel} for every action at once when $K > 2$: summed over the $K$ options, the right-hand side is $1 + (K - 2)f$, not 1. The channel identity is therefore a statement about the logged action. Uncorrected, its ECE against clean correctness is .059, .164 and .214 at f 0.1, 0.2 and 0.3, against the clean run's .026. Backward correction inverts \eqref{eq:channel} on the top-1 probability:
\begin{equation}
  \label{eq:backward}
  \hat p = \frac{p - f}{1 - 2f},
\end{equation}
which amplifies any error in $p$ by $1/(1 - 2f)$, at 1.25, 1.67 and 2.5. A corrected value outside $[0, 1]$ is clipped to $[0, 1]$, and the corrected ECE bins the clipped top-1 value alone, with no renormalisation over the other options. It brings ECE to .020, .051 and .068 (Table~\ref{tab:A4}). The accuracy loss of CAPS grows with f: $-$.006, $-$.022 and $-$.041. Against positive-only SFT's loss it is smaller at 0.2, +.017 [+.007, +.028], touches zero at 0.1, +.009 [+.000, +.018], and is level at 0.3, $-$.003 [$-$.014, +.007]. The two learners face different corruptions. Positive-only SFT trains on mislabelled positives, about .13 of the positives in the log at f 0.3, computed from the stored outcome rate, flipped outcome rate and flip share. CAPS sees symmetric outcome noise. Noise softens positive-only SFT's calibration, from ECE .078 clean to .043 to .050. Trained to the end, CAPS memorises the single flip draw and falls to .734 at 0.3, below zero-shot's .851. Early stopping on the held-out log score of logged outcomes \eqref{eq:select} picked steps before the collapse in all three runs. A soft-target diagnostic (Appendix~\ref{sec:A4}) trains on the exact per-action target and so has no draw variance left; its residual, channel-scale ECE .019 to .024 against the reference .014, bounds from above the cost of the misspecification noted in Section~\ref{sec:3}, and the rest of the excess is single-draw variance.

\textbf{Takeaway.} Under flips at a known rate, one seed per rate, CAPS needs backward correction to recover most of the clean calibration and early stopping on held-out logged outcomes to avoid memorising the draw.

\begin{figure}[t]
\centering
\includegraphics[width=\textwidth]{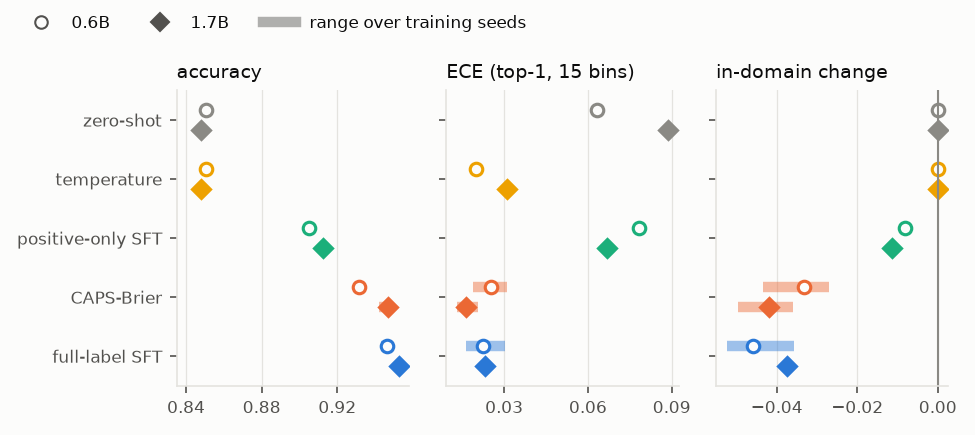}
\caption{Scale on the Banking77 log at N 5000: 0.6B (open circles) against 1.7B (filled diamonds) for every learner. Left: accuracy on the Banking77 test split; centre: its ECE; right: in-domain (CLINC150) test accuracy minus zero-shot. Learners run down the vertical axis. CAPS-Brier at both sizes and full-label SFT at 0.6B are plotted as the mean of three training seeds, with the range over those seeds as a thick bar. Every other marker is seed 0, the only seed. No record-bootstrap interval is drawn. Provenance: Table~\ref{tab:prov}, F\ref{fig:v4_scale}.}\label{fig:v4_scale}
\end{figure}

\textbf{Scale.} At 1.7B, with its own log, CAPS-Brier on three training seeds reaches accuracy .947 (.942, .949). Its seed mean leads positive-only SFT by +.034 [+.026, +.041], a three-seed mean against one baseline seed. Against full-label SFT's one seed it is level on seed 0, $-$.004 [$-$.010, +.002], and .6 points below as a seed mean, $-$.006 [$-$.011, $-$.002]. At 0.6B the gap is 1.5 points [1.0, 2.0], between three-seed means (Figure~\ref{fig:v4_scale}; Table~\ref{tab:A5b}). Every 1.7B CAPS-Brier seed is above every 0.6B one. On these point estimates, the larger model therefore helps the learner from logged bandit feedback most.

\textbf{Takeaway.} At 1.7B, against one full-label seed, CAPS-Brier is level on seed 0 and .6 points below as a seed mean, and every 1.7B seed is above every 0.6B seed.

\textbf{Forgetting and selective prediction.} At N 5000, CAPS-Brier costs 2.7 to 4.3 points of in-domain accuracy across seeds, full-label SFT 3.6 to 5.2 and positive-only SFT .8. CAPS-Brier's in-domain ECE stays within .010 of zero-shot's (Table~\ref{tab:A1c}), and every trained learner gains 1.7 to 2.7 points on the unseen intents. SST-5, the other training domain, was not evaluated after adaptation. At a fixed confidence threshold of .95, positive-only SFT keeps .916 of answers at .942 accuracy, so it passes on about 6 percent errors as confident. CAPS-Brier keeps .743 at .986 (Table~\ref{tab:A1c}, with no stored interval). At a fixed error budget of .05, CAPS-Brier keeps .957 of answers against positive-only SFT's .888. On the 500-record Banking77 subset, gpt-4.1-mini scores .918. Paired over the same records, CAPS-Brier's seed mean is +.014 [$-$.010, +.037] above it, within noise (per seed, +.008 to +.018). Full-label SFT's is +.029 [+.003, +.053] (per seed, +.026 to +.032).

\textbf{Takeaway.} CAPS-Brier forgets about as much in-domain accuracy as full-label SFT and more than positive-only SFT, keeps in-domain ECE within .010 of zero-shot, and at a .95 threshold passes fewer confident errors than positive-only SFT, with no interval stored for that figure.

\subsection{A second domain: emotion}\label{sec:4.4}

\begin{figure}[t]
\centering
\includegraphics[width=\textwidth]{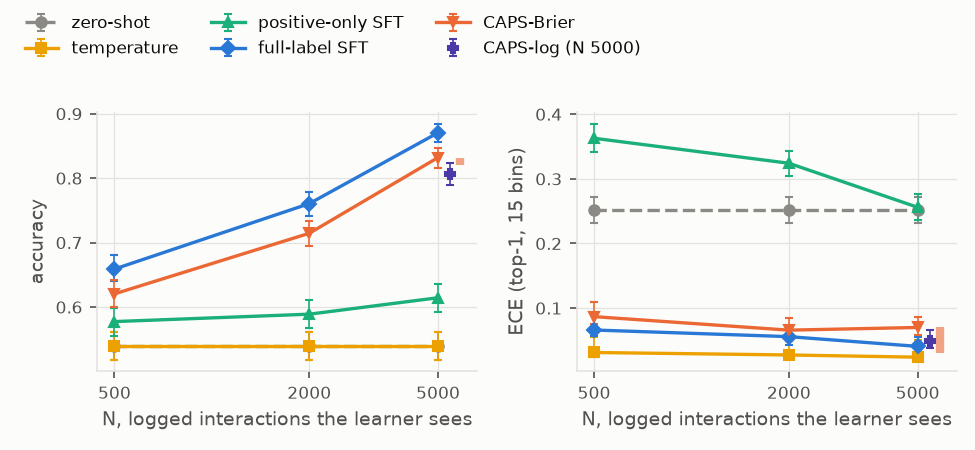}
\caption{Emotion N-curve, 0.6B, on the emotion test split: every learner at seed 0 with 95 percent record-bootstrap intervals, CAPS-log at N 5000, and CAPS-Brier's range over three training seeds. Provenance: Table~\ref{tab:prov}, F\ref{fig:v4_emotion_n_curve}.}\label{fig:v4_emotion_n_curve}
\end{figure}

The emotion log domain is the dair-ai/emotion train split, 15958 messages in a fixed seeded shuffle, each a choice over all 6 labels with no \texttt{other} option. The same 0.6B SFT head logs it. The logger's top choice is right .542 of the time and its sampled action .474, so 8393 of 15958 outcomes are negative. That share, .526, compares with $1 - .739 = .261$ on Banking77. Evaluation is on the full emotion test split, 2000 records, and Table~\ref{tab:main} gives N 5000. Positive-only SFT and full-label SFT have one seed here. CAPS-Brier leads positive-only SFT at every N on seed 0, by +.043 [+.027, +.060], +.126 [+.109, +.144] and +.217 [+.197, +.238] in accuracy at N 500, 2000 and 5000. Its ECE is lower by .186 to .276 (Figure~\ref{fig:v4_emotion_n_curve}; Table~\ref{tab:A7}). Over three seeds at N 5000, its seed mean is $-$.043 [$-$.056, $-$.031] below full-label SFT in accuracy, a three-seed mean against one baseline seed. In ECE it is +.006 [$-$.005, +.018], unresolved at three seeds (per-seed ranges in Table~\ref{tab:A7d}). The per-N details, CAPS-log, the temperature learner, forgetting, the label ceiling and the API reference are in Appendix~\ref{sec:A7}.

As on Banking77, CAPS beats positive-only SFT in accuracy and ECE from N 2000, and here in accuracy from N 500 too, where Banking77 tied. The gap to full-label SFT at N 5000 is larger than on Banking77: $-$.039 [$-$.052, $-$.027] against $-$.018 [$-$.025, $-$.011] on seed 0. For the seed means it is $-$.043 against $-$.015, the emotion one against one full-label SFT seed and the Banking77 one between three-seed means. The task is harder and the logging policy wrong more often, but the two change together, so the cause is not isolated. The lead over positive-only SFT is nine times Banking77's at N 5000: with fewer than half the outcomes positive, discarding the negatives discards most of the log.

\textbf{Takeaway.} On emotion, where the logger is wrong more often than right, discarding the negatives discards most of the log, and CAPS leads positive-only SFT by 21.7 points on one seed. Its gap to full-label SFT is larger than on Banking77, and the cause is not isolated.

\section{Limitations}\label{sec:5}

\begin{itemize}
  \item \textbf{Simulated feedback.} Outcomes come from gold labels, cleanly or flipped symmetrically at a known rate. Real feedback is delayed and biased towards some errors, and its noise rate, which backward correction needs, is unknown. In clean simulation a positive reveals the gold label, so positive-only SFT holds full labels for the interactions it keeps, and CAPS adds one exclusion per negative. Real success signals need not identify the label. Positive-only SFT with 3342 labels at N 5000 (.905) is below full-label SFT at N 2000 with 1800 (.924) (Table~\ref{tab:A1}). This comparison points to the selection rather than the label count, though the two subsets also differ in size and passes. Under flips the two learners face different corruptions, mislabelled positives for positive-only SFT and symmetric outcome noise for CAPS. The compression towards the channel is global, with in-domain ECE up by .050 for the run at f 0.2 (Table~\ref{tab:A1c}).
  \item \textbf{Domains, loggers and seeds.} There are two adaptation domains with one logging policy per domain, the starting model itself at epsilon 0.1, whose top choice is right .836 of the time on Banking77, .542 on emotion and .837 at 1.7B. A second log draw on Banking77 is a robustness check. A logger other than the starting model, or another epsilon, would change what the positives contain. Positive-only SFT has one training seed everywhere. Full-label SFT has three at N 5000 at 0.6B on Banking77 and one on emotion and at 1.7B. At N 500 and 2000 every learner has one, and the noise sweep has one seed per rate. Every interval is a record bootstrap without training-seed variance. Differences against a one-seed baseline are therefore seed means against one baseline seed, with the per-seed range beside them. On Banking77, \texttt{other} is always wrong, a free negative and a learnable rule; emotion has no \texttt{other} option.
  \item \textbf{Equal steps.} All trained learners take the same number of steps, so positive-only SFT makes more passes over its smaller set (13.4 epochs at N 5000 on Banking77 against 10.0). Selection limits this, and the selected steps are in Table~\ref{tab:A1c}, but part of its miscalibration may be budget.
  \item \textbf{Missing off-policy arms.} The weighted arms reweight the direct-method loss, and the IPS arms clip 12 to 12.5 percent of their weights over the training prefix. There is no counterfactual-risk policy learner (POEM or BanditNet; \citealp{swaminathan2015crm,joachims2018banditnet}), no doubly robust arm \citep{dudik2011doubly} and no self-normalised estimate of the learned policy's value. The no-gain result is therefore shown against reweighted regression on this log, not against the methods built around the weights.
  \item \textbf{Label noise, judged by two language models, no human.} The judges were Claude (Fable 5.1 and Opus 5.5, via Claude Code) and GPT-Astra; the Banking77 pass was committed on 4 October 2026 and the emotion pass on 15 October 2026 (Appendix~\ref{sec:setup}). Exact model identifiers were not recorded. They find a label problem in .12 and .14 of a uniform Banking77 sample and in .32 and .40 of an emotion sample. The clean feedback inherits these labels, so every calibration number here is measured against labels that are themselves noisy. The natural follow-up is a synthetic environment in which the true outcome probability $p^*$ is known. There the error $\sum (p - p^*)^2$ can be measured directly, and Bernoulli($p^*$) feedback is the truth rather than a corruption.
  \item \textbf{Selection.} Checkpoints are every 50 steps. At N 2000, where the log-score arms peaked at the first candidate, a finer grid could favour them, and at N 500 only 50 interactions select.
  \item \textbf{Not run.} Further positive-only SFT seeds; baseline seeds on emotion and at 1.7B; a second logger or epsilon; counterfactual-risk and doubly robust arms; seeds for the noise sweep; a KTO-style objective \citep{ethayarajh2024kto} on the readout.
  \item \textbf{Coverage of the claim.} The gap to full labels does not shrink with N, so the claim is that using failures beats discarding them, not that it replaces labels.
\end{itemize}

\section{Conclusion}\label{sec:6}

A decision head that can only see whether its chosen action was right should be fitted as the direct method with a proper score. The score compares the chosen action's probability with its own logged outcome, by the pathwise gradient. That is CAPS; with the log score it is positive-only fine-tuning plus an unlikelihood term on every refuted option. From 2000 interactions it beats positive-only fine-tuning, which discards the negatives, in accuracy, calibration and AURC on both domains, against one positive-only training seed (Sections~\ref{sec:4.3} and~\ref{sec:4.4}). On this log propensity weights add nothing (Section~\ref{sec:4.3}). Under feedback flipped at a known rate, the chosen action's probability tracks the channel rate on average, and backward correction recovers most of the clean calibration. Early stopping on the held-out log score of logged outcomes stopped it before it memorised the single noisy draw in all three runs (Section~\ref{sec:4.3}). CAPS stays below full-label training on both domains (Sections~\ref{sec:4.3} and~\ref{sec:4.4}). On Banking77 at N 5000 the gap is 1.5 points at 0.6B, between three-seed means, and at 1.7B it is level to .6 points, a three-seed mean against one full-label seed. Failures are therefore worth keeping, not a substitute for labels.

The two negative results frame it. Scoring the sampled action's own probability with the Brier or log score as a detached REINFORCE reward rewards wrong actions (Sections~\ref{sec:3} and~\ref{sec:4.2}). This is a corollary of the general rule that detaching a parameter-dependent reward drops the pathwise term. It is distinct from proper-score rewards on a separately emitted confidence. With full labels, consistent with OpenJev-RLCD's direct-scorer ablation, none of the four bandit arms tested beats cross-entropy plus one temperature on mean ECE over the four unseen label schemas (Section~\ref{sec:4.2}).

\section{Reproducibility}\label{sec:7}

Code and data are at \url{https://github.com/yuan-phd/jevmark} (Apache-2.0), with tags v1, v2, v3 and v3-final for the project's stages, v4-paper for the first submission and v5-paper for the state this paper reports. The adapters of the 0.6B and 1.7B SFT heads and of CAPS-Brier adapted to Banking77 are on the Hugging Face Hub (\texttt{yuanphd/jevmark}), under the names given in Appendix~\ref{sec:B}. The deployment logs and every run's per-record results are the Hugging Face dataset \texttt{yuanphd/jevmark-runs} (\url{https://huggingface.co/datasets/yuanphd/jevmark-runs}). The logs at flip rates 0.1 and 0.3 are rebuilt exactly from the main log by a script in the repository (Appendix~\ref{sec:B}). Every run's metrics file records its commit, dirty flag, data sha256 and adapter sha256. Every table and figure is regenerated on CPU from the committed metrics files and these run files (Appendix~\ref{sec:B}). The paired comparison with gpt-4.1-mini also needs its replies, which are not published; their sha256 is recorded.

\section{Use of AI assistance}\label{sec:8}

The code, the experiment pipeline, the project reports and the first draft of this paper were written with Claude Code (Anthropic) under the author's direction and review. The label audits were judged by two language models, Claude (Fable 5.1 and Opus 5.5, via Claude Code) and GPT-Astra, with the Banking77 pass committed on 4 October 2026 and the emotion pass on 15 October 2026, and no human pass. Exact model identifiers were not recorded. The limitations (Section~\ref{sec:5}) and Appendix~\ref{sec:setup} say the same. The related-work survey was compiled with an AI research tool, and every bibliographic entry was then verified against primary sources. The four external reviews used during revision were also produced by a language model. All experiments, numbers, claims and the final text are the author's responsibility.

\bibliographystyle{plainnat}
\bibliography{references}

\clearpage
\appendix
\counterwithin{table}{subsection}
\section{Full tables}\label{sec:A}

Every bracketed interval in this appendix is a 95 percent record bootstrap, paired across runs where it is a difference. It resamples test records and carries no training-seed variance. A seed mean set against a learner with one seed is a seed mean against one baseline seed.

\subsection{The N-curve on the Banking77 test split, 0.6B, seed 0}\label{sec:A1}

\begin{table}[H]
\centering
\footnotesize
\setlength{\tabcolsep}{3pt}
\caption{The N-curve on the Banking77 test split, 0.6B, seed 0. Brackets are 95 percent record-bootstrap intervals; ECE intervals sit mostly above the point estimate because binned ECE is biased upwards under resampling. AURC and Cov@.05 rank by top-1 probability.}
\label{tab:A1}
\begin{tabular}{llrrrrr}
\toprule
N & metric & zero-shot & temperature & positive-only SFT & full-label SFT & CAPS-Brier \\
\midrule
500 & accuracy & .851 [.839, .863] & .851 & .886 [.874, .896] & .904 [.894, .914] & .888 [.877, .898] \\
2000 & accuracy & .851 & .851 & .890 [.879, .901] & .924 [.915, .933] & .917 [.907, .926] \\
5000 & accuracy & .851 & .851 & .905 [.896, .915] & .947 [.938, .955] & .929 [.920, .938] \\
500 & ECE & .063 [.054, .074] & .023 [.021, .036] & .091 [.082, .102] & .043 [.035, .052] & .031 [.026, .044] \\
2000 & ECE & .063 & .018 [.017, .032] & .082 [.072, .092] & .017 [.013, .027] & .022 [.017, .031] \\
5000 & ECE & .063 & .020 [.016, .032] & .078 [.069, .087] & .030 [.024, .038] & .026 [.020, .034] \\
500 & Brier & .229 [.211, .247] & .219 [.204, .234] & .200 [.182, .220] & .149 [.133, .164] & .173 [.159, .188] \\
2000 & Brier & .229 & .219 [.204, .234] & .185 [.167, .203] & .114 [.102, .126] & .127 [.113, .140] \\
5000 & Brier & .229 & .219 [.203, .234] & .169 [.152, .186] & .082 [.070, .094] & .111 [.098, .124] \\
500 & NLL & .582 [.531, .633] & .518 [.483, .555] & .764 [.683, .852] & .339 [.301, .378] & .392 [.354, .430] \\
2000 & NLL & .582 & .518 [.481, .554] & .622 [.551, .692] & .236 [.209, .266] & .299 [.264, .332] \\
5000 & NLL & .582 & .517 [.479, .555] & .663 [.581, .739] & .212 [.175, .249] & .274 [.239, .311] \\
500 & AURC & .0333 [.0283, .0392] & .0327 [.0278, .0387] & .0253 [.0206, .0305] & .0179 [.0140, .0223] & .0218 [.0179, .0266] \\
2000 & AURC & .0333 & .0328 [.0278, .0387] & .0230 [.0186, .0276] & .0106 [.0084, .0130] & .0132 [.0104, .0162] \\
5000 & AURC & .0333 & .0328 [.0279, .0387] & .0202 [.0164, .0241] & .0063 [.0045, .0082] & .0110 [.0086, .0135] \\
500 & Cov@.05 & .756 [.726, .793] & .771 [.743, .794] & .853 [.814, .881] & .884 [.857, .916] & .836 [.815, .865] \\
2000 & Cov@.05 & .756 & .771 [.742, .794] & .857 [.819, .888] & .946 [.913, .965] & .922 [.900, .946] \\
5000 & Cov@.05 & .756 & .765 [.742, .794] & .888 [.855, .923] & .996 [.982, 1.000] & .957 [.938, .974] \\
\bottomrule
\end{tabular}
\end{table}

Cov@.10 at N 500 is .895, .901, .972, 1.000 and .974 in column order. At N 2000 it is 1 for every trained learner except positive-only SFT (.985), and at N 5000 for every trained learner; Cov@.20 is 1 throughout.

CAPS-Brier closes 69, 91 and 81 percent of the accuracy gap between zero-shot and full-label SFT at N 500, 2000 and 5000, computed from the stored accuracies. One temperature is monotone within a question but can reorder top-1 probabilities across questions. At N 5000 the temperature learner moves Cov@.05 from .756 to .765, and on CAPS-Brier, whose fitted temperature is near 1, it moves AURC by under .0001.

\begin{landscape}
\begin{table}[H]
\centering
\footnotesize
\setlength{\tabcolsep}{3pt}
\caption{The N-curve on the Banking77 test split, 0.6B, seed 0: paired differences, CAPS-Brier minus the other learner.}
\label{tab:A1b}
\begin{tabular}{llrrrrrr}
\toprule
N & vs & accuracy & ECE & Brier & NLL & AURC & Cov@.05 \\
\midrule
500 & positive-only SFT & +.002 [$-$.006, +.009] & $-$.060 [$-$.066, $-$.048] & $-$.027 [$-$.037, $-$.018] & $-$.373 [$-$.426, $-$.319] & $-$.0035 [$-$.0066, $-$.0007] & $-$.016 [$-$.027, +.014] \\
2000 & positive-only SFT & +.027 [+.019, +.035] & $-$.060 [$-$.067, $-$.050] & $-$.058 [$-$.069, $-$.048] & $-$.323 [$-$.373, $-$.276] & $-$.0099 [$-$.0131, $-$.0070] & +.065 [+.048, +.093] \\
5000 & positive-only SFT & +.024 [+.016, +.031] & $-$.052 [$-$.060, $-$.043] & $-$.058 [$-$.069, $-$.046] & $-$.389 [$-$.443, $-$.333] & $-$.0092 [$-$.0118, $-$.0068] & +.069 [+.046, +.094] \\
500 & full-label SFT & $-$.017 [$-$.023, $-$.011] & $-$.011 [$-$.016, $-$.001] & +.024 [+.018, +.030] & +.052 [+.036, +.070] & +.0038 [+.0018, +.0058] & $-$.048 [$-$.063, $-$.028] \\
2000 & full-label SFT & $-$.007 [$-$.014, +.000] & +.005 [$-$.002, +.011] & +.013 [+.007, +.019] & +.062 [+.046, +.078] & +.0026 [+.0011, +.0043] & $-$.024 [$-$.035, $-$.002] \\
5000 & full-label SFT & $-$.018 [$-$.025, $-$.011] & $-$.004 [$-$.010, +.003] & +.029 [+.020, +.038] & +.062 [+.037, +.086] & +.0047 [+.0031, +.0063] & $-$.039 [$-$.051, $-$.024] \\
\bottomrule
\end{tabular}
\end{table}

\begin{table}[H]
\centering
\footnotesize
\setlength{\tabcolsep}{3pt}
\caption{Forgetting, selected steps and the .95 threshold at N 5000, 0.6B. Forgetting is each run minus zero-shot (first row, absolute) on the in-domain test split (CLINC150, 11800 questions) and the unseen-intents test split (6000). The selected step is out of 1407, with the epochs over the interactions the learner trains on. The last column is the share of Banking77 test answers kept at a top-1 threshold of .95 and their accuracy; no interval is stored for it.}
\label{tab:A1c}
\begin{tabular}{lrrrrrr}
\toprule
run & in-domain accuracy & in-domain ECE & unseen-intents accuracy & unseen-intents ECE & selected step (epochs) & kept at .95 (accuracy) \\
\midrule
zero-shot & .956 [.952, .960] & .013 [.011, .017] & .892 [.883, .900] & .068 [.062, .077] &  & .607 (.978) \\
temperature & +.000 [+.000, +.000] & $-$.000 [$-$.005, +.005] & +.000 [+.000, +.000] & $-$.026 [$-$.029, $-$.024] &  & .189 (.998) \\
positive-only SFT & $-$.008 [$-$.011, $-$.006] & +.017 [+.014, +.019] & +.020 [+.015, +.025] & +.001 [$-$.004, +.006] & 200 (1.9 of 13.4) & .916 (.942) \\
IPS positive-only SFT & $-$.009 [$-$.011, $-$.006] & +.017 [+.014, +.019] & +.023 [+.017, +.028] & $-$.001 [$-$.007, +.004] & 200 (1.9 of 13.4) & .917 (.941) \\
CAPS-Brier seed 0 & $-$.027 [$-$.031, $-$.023] & $-$.001 [$-$.004, +.004] & +.025 [+.018, +.032] & $-$.031 [$-$.037, $-$.023] & 300 (2.1 of 10.0) & .743 (.986) \\
CAPS-Brier seed 1 & $-$.043 [$-$.048, $-$.039] & +.010 [+.006, +.015] & +.020 [+.012, +.027] & $-$.026 [$-$.033, $-$.018] & 550 (3.9 of 10.0) & .794 (.984) \\
CAPS-Brier seed 2 & $-$.029 [$-$.033, $-$.025] & +.001 [$-$.002, +.005] & +.026 [+.019, +.033] & $-$.032 [$-$.039, $-$.024] & 250 (1.8 of 10.0) & .704 (.989) \\
CAPS-Brier, second log draw & $-$.030 [$-$.034, $-$.026] & +.008 [+.003, +.013] & +.025 [+.018, +.032] & $-$.034 [$-$.039, $-$.026] & 350 (2.5 of 10.0) & .785 (.986) \\
CAPS-Brier, f 0.2 & $-$.014 [$-$.017, $-$.011] & +.050 [+.042, +.056] & +.018 [+.012, +.025] & $-$.018 [$-$.029, $-$.008] & 250 (1.8 of 10.0) & .000 \\
CAPS-log seed 0 & $-$.038 [$-$.043, $-$.034] & +.007 [+.003, +.011] & +.024 [+.016, +.031] & $-$.028 [$-$.034, $-$.020] & 250 (1.8 of 10.0) & .762 (.987) \\
CAPS-log seed 1 & $-$.046 [$-$.051, $-$.042] & +.007 [+.004, +.012] & +.025 [+.018, +.033] & $-$.028 [$-$.035, $-$.020] & 250 (1.8 of 10.0) & .780 (.988) \\
CAPS-log seed 2 & $-$.043 [$-$.047, $-$.038] & +.007 [+.003, +.012] & +.017 [+.010, +.024] & $-$.025 [$-$.033, $-$.017] & 250 (1.8 of 10.0) & .654 (.989) \\
IPS CAPS-log & $-$.036 [$-$.040, $-$.032] & +.007 [+.003, +.011] & +.021 [+.013, +.028] & $-$.024 [$-$.030, $-$.016] & 250 (1.8 of 10.0) & .729 (.988) \\
REINFORCE, reward $r - p$ & $-$.026 [$-$.030, $-$.022] & +.007 [+.004, +.010] & +.022 [+.015, +.029] & $-$.023 [$-$.030, $-$.015] & 300 (2.1 of 10.0) & .739 (.984) \\
full-label SFT seed 0 & $-$.052 [$-$.057, $-$.047] & +.043 [+.038, +.047] & +.022 [+.014, +.030] & $-$.007 [$-$.015, $-$.000] & 300 (2.1 of 10.0) & .888 (.982) \\
full-label SFT seed 1 & $-$.036 [$-$.040, $-$.032] & +.015 [+.011, +.018] & +.027 [+.019, +.034] & $-$.026 [$-$.034, $-$.019] & 200 (1.4 of 10.0) & .785 (.990) \\
full-label SFT seed 2 & $-$.050 [$-$.054, $-$.045] & +.025 [+.020, +.029] & +.020 [+.012, +.027] & $-$.020 [$-$.027, $-$.013] & 250 (1.8 of 10.0) & .786 (.990) \\
\bottomrule
\end{tabular}
\end{table}
\end{landscape}

\subsection{Training seeds at N 5000}\label{sec:A2}

\begin{landscape}
\begin{table}[H]
\centering
\footnotesize
\setlength{\tabcolsep}{2pt}
\caption{Training seeds at N 5000: mean (range) over three training seeds, and CAPS-Brier's second log draw (seed 0 on the log drawn with seed 1). CAPS-Brier's NLL is given to four decimals: its seeds are .2737, .2742 and .2737, a spread of .0005.}
\label{tab:A2}
\begin{tabular}{llrrrrr}
\toprule
learner & accuracy & ECE & Brier & NLL & AURC & Cov@.05 \\
\midrule
CAPS-Brier & .931 (.929, .935) & .025 (.019, .031) & .109 (.105, .111) & .2739 (.2737, .2742) & .0107 (.0104, .0110) & .962 (.957, .971) \\
CAPS-log & .933 (.931, .934) & .022 (.019, .027) & .105 (.101, .110) & .254 (.239, .275) & .0099 (.0089, .0113) & .966 (.959, .970) \\
full-label SFT & .946 (.943, .949) & .022 (.016, .030) & .082 (.079, .086) & .192 (.175, .212) & .0067 (.0063, .0071) & .994 (.986, .999) \\
CAPS-Brier, second log draw & .934 [.925, .943] & .026 [.022, .034] & .103 [.090, .115] & .256 [.221, .293] & .0102 [.0076, .0128] & .969 [.949, .989] \\
\bottomrule
\end{tabular}
\end{table}

\begin{table}[H]
\centering
\footnotesize
\setlength{\tabcolsep}{3pt}
\caption{Training seeds at N 5000: seed-mean differences. The first is a seed mean against positive-only SFT's one seed, with the range of the per-seed differences below it. The other two are between three-seed means, with a verdict below from the seed ranges in Table~\ref{tab:A2}: where they overlap, the difference is unresolved at three seeds.}
\label{tab:A2b}
\begin{tabular}{llrrrrr}
\toprule
seed mean minus & accuracy & ECE & Brier & NLL & AURC & Cov@.05 \\
\midrule
CAPS-Brier minus positive-only SFT & +.026 [+.019, +.033] & $-$.053 [$-$.059, $-$.044] & $-$.060 [$-$.071, $-$.048] & $-$.389 [$-$.443, $-$.333] & $-$.0095 [$-$.0121, $-$.0070] & +.074 [+.052, +.098] \\
\quad per-seed range & (+.024, +.030) & ($-$.059, $-$.048) & ($-$.064, $-$.058) & ($-$.3891, $-$.3886) & ($-$.0097, $-$.0092) & (+.069, +.083) \\
CAPS-Brier minus full-label SFT & $-$.015 [$-$.020, $-$.010] & +.003 [$-$.001, +.008] & +.027 [+.020, +.034] & +.082 [+.061, +.101] & +.0040 [+.0026, +.0053] & $-$.032 [$-$.041, $-$.019] \\
\quad seed ranges & apart & overlap: unresolved & apart & apart & apart & apart \\
CAPS-log minus CAPS-Brier & +.002 [$-$.001, +.005] & $-$.003 [$-$.006, +.000] & $-$.005 [$-$.008, $-$.001] & $-$.020 [$-$.029, $-$.010] & $-$.0008 [$-$.0016, +.0000] & +.004 [$-$.001, +.010] \\
\quad seed ranges & overlap: unresolved & overlap: unresolved & overlap: unresolved & overlap: unresolved & overlap: unresolved & overlap: unresolved \\
\bottomrule
\end{tabular}
\end{table}
\end{landscape}

Training seeds and the second log draw move CAPS-Brier's accuracy by at most .6 points. A temperature fitted on CAPS-Brier's own training interactions comes out at .951 to 1.039 and moves ECE by at most .003.

\subsection{Off-policy baselines, seed 0}\label{sec:A3}

\begin{landscape}
\begin{table}[H]
\centering
\footnotesize
\setlength{\tabcolsep}{3pt}
\caption{Off-policy baselines, seed 0, on the Banking77 test split. The last column counts the clipped weights. For IPS CAPS-log it counts over the training prefix, and for IPS positive-only SFT over the positives it trains on (over the prefix it is the same as IPS CAPS-log's). For the logged REINFORCE arm with reward $r - p$ it counts over all training steps.}
\label{tab:A3}
\begin{tabular}{llrrrrrrrr}
\toprule
N & learner & accuracy & ECE & Brier & NLL & AURC & Cov@.05 & accuracy minus CAPS-Brier & weights clipped \\
\midrule
2000 & positive-only SFT & .890 [.879, .901] & .082 [.072, .092] & .185 [.167, .203] & .622 [.551, .692] & .0230 [.0186, .0276] & .857 [.819, .888] &  &  \\
2000 & IPS positive-only SFT & .892 [.881, .902] & .077 [.068, .087] & .183 [.165, .200] & .593 [.525, .659] & .0233 [.0189, .0282] & .856 [.819, .891] & $-$.025 [$-$.032, $-$.018] & 14 of 1332 \\
2000 & CAPS-log & .907 [.897, .917] & .014 [.012, .025] & .140 [.128, .153] & .319 [.289, .351] & .0161 [.0132, .0194] & .892 [.865, .926] & $-$.010 [$-$.016, $-$.004] &  \\
2000 & IPS CAPS-log & .906 [.897, .916] & .012 [.010, .024] & .143 [.129, .156] & .323 [.291, .357] & .0168 [.0137, .0202] & .892 [.858, .916] & $-$.011 [$-$.016, $-$.005] & 216 of 1800 \\
2000 & REINFORCE, reward $r - p$ & .916 [.906, .925] & .030 [.024, .040] & .132 [.118, .146] & .325 [.287, .365] & .0147 [.0116, .0180] & .919 [.894, .939] & $-$.001 [$-$.006, +.003] & 132 of 17800 \\
2000 & CAPS-Brier & .917 [.907, .926] & .022 [.017, .031] & .127 [.113, .140] & .299 [.264, .332] & .0132 [.0104, .0162] & .922 [.900, .946] &  &  \\
2000 & full-label SFT & .924 [.915, .933] & .017 [.013, .027] & .114 [.102, .126] & .236 [.209, .266] & .0106 [.0084, .0130] & .946 [.913, .965] &  &  \\
5000 & positive-only SFT & .905 [.896, .915] & .078 [.069, .087] & .169 [.152, .186] & .663 [.581, .739] & .0202 [.0164, .0241] & .888 [.855, .923] &  &  \\
5000 & IPS positive-only SFT & .905 [.895, .915] & .078 [.069, .087] & .169 [.151, .186] & .650 [.568, .723] & .0204 [.0165, .0246] & .891 [.864, .916] & $-$.024 [$-$.031, $-$.016] & 25 of 3342 \\
5000 & CAPS-log & .933 [.925, .942] & .022 [.018, .031] & .103 [.091, .116] & .248 [.214, .282] & .0096 [.0073, .0120] & .969 [.949, .986] & +.005 [$-$.000, +.010] &  \\
5000 & IPS CAPS-log & .930 [.921, .939] & .026 [.020, .034] & .106 [.093, .118] & .253 [.221, .287] & .0097 [.0075, .0120] & .963 [.944, .981] & +.001 [$-$.004, +.007] & 561 of 4500 \\
5000 & REINFORCE, reward $r - p$ & .928 [.918, .937] & .029 [.024, .039] & .115 [.102, .129] & .293 [.255, .333] & .0121 [.0096, .0149] & .953 [.924, .970] & $-$.001 [$-$.006, +.003] & 240 of 44916 \\
5000 & CAPS-Brier & .929 [.920, .938] & .026 [.020, .034] & .111 [.098, .124] & .274 [.239, .311] & .0110 [.0086, .0135] & .957 [.938, .974] &  &  \\
5000 & full-label SFT & .947 [.938, .955] & .030 [.024, .038] & .082 [.070, .094] & .212 [.175, .249] & .0063 [.0045, .0082] & .996 [.982, 1.000] &  &  \\
\bottomrule
\end{tabular}
\end{table}
\end{landscape}

Every arm that keeps the negatives leads positive-only SFT, by +.016 to +.028 in accuracy. IPS positive-only SFT equals positive-only SFT: +.002 [$-$.001, +.005] in accuracy at N 2000 and +.000 [$-$.003, +.004] at N 5000. In AURC the differences are +.0003 [$-$.0003, +.0009] and +.0002 [$-$.0005, +.0010]. At N 5000, IPS CAPS-log is ahead of CAPS-Brier on Brier and NLL (.106 against .111, .253 against .274), and the unweighted CAPS-log is further ahead (.103, .248). These single-seed differences are smaller than CAPS-log's own seed range (.101 to .110 in Brier, .239 to .275 in NLL). Whether the gain is the log score's or the weight's is therefore unresolved at three seeds. CAPS-log's in-domain forgetting ($-$.038 to $-$.046 across seeds) overlaps CAPS-Brier's ($-$.027 to $-$.043; Table~\ref{tab:A1c}), also unresolved at three seeds. CAPS-log is behind CAPS-Brier at N 2000 on one seed ($-$.010 [$-$.016, $-$.004]).

\begin{figure}[t]
\centering
\includegraphics[width=\textwidth]{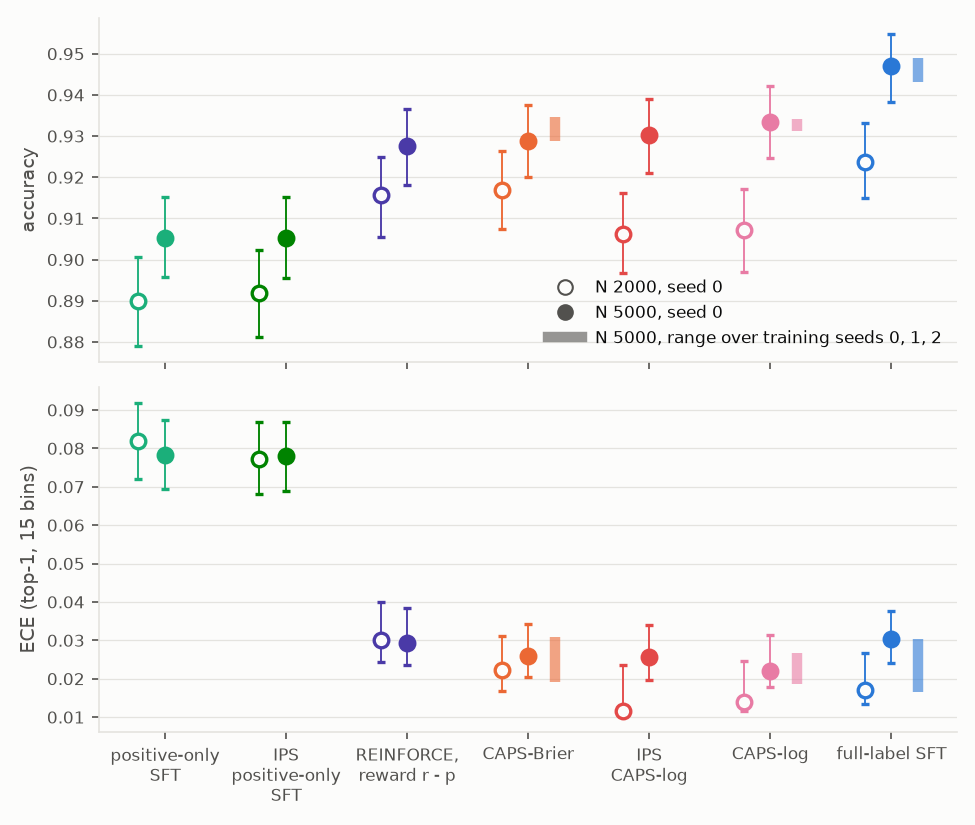}
\caption{Off-policy baselines on the Banking77 log, 0.6B: accuracy (top) and ECE (bottom) on the Banking77 test split, 3080 records. Open circles are N 2000 and filled circles N 5000, both seed 0, and their thin whiskers are 95 percent record-bootstrap intervals. The thick bars are the range over training seeds 0, 1 and 2 at N 5000 for the three arms with three seeds, CAPS-Brier, CAPS-log and full-label SFT. Provenance: Table~\ref{tab:prov}, F\ref{fig:v4_offpolicy}.}\label{fig:v4_offpolicy}
\end{figure}

\subsection{Noise-rate sweep, N 5000, seed 0}\label{sec:A4}

\begin{table}[H]
\centering
\footnotesize
\setlength{\tabcolsep}{2.5pt}
\caption{Noise-rate sweep, N 5000, seed 0, one seed per rate: accuracy and calibration against clean correctness. ``Minus clean'' is the paired accuracy loss from the clean run. The backward-corrected ECE scores $(p - f)/(1 - 2f)$ on the top-1 probability, clipped to $[0, 1]$. The last column is CAPS-Brier's accuracy loss minus positive-only SFT's at the same rate.}
\label{tab:A4}
\begin{tabular}{llrrrrr}
\toprule
f & learner & accuracy & minus clean & ECE, uncorrected & ECE, backward-corrected & loss minus positive-only SFT's \\
\midrule
0 & CAPS-Brier & .929 [.920, .938] &  & .026 [.020, .034] &  &  \\
0.1 & CAPS-Brier & .923 [.914, .932] & $-$.006 [$-$.013, +.001] & .059 [.055, .069] & .020 [.017, .030] & +.009 [+.000, +.018] \\
0.2 & CAPS-Brier & .907 [.897, .916] & $-$.022 [$-$.030, $-$.014] & .164 [.157, .173] & .051 [.042, .059] & +.017 [+.007, +.028] \\
0.3 & CAPS-Brier & .888 [.877, .898] & $-$.041 [$-$.050, $-$.032] & .214 [.206, .225] & .068 [.058, .078] & $-$.003 [$-$.014, +.007] \\
0 & positive-only SFT & .905 [.896, .915] &  & .078 [.069, .087] &  &  \\
0.1 & positive-only SFT & .890 [.880, .901] & $-$.015 [$-$.021, $-$.008] & .046 [.039, .057] & .071 [.062, .082] &  \\
0.2 & positive-only SFT & .866 [.855, .878] & $-$.039 [$-$.048, $-$.031] & .050 [.040, .060] & .105 [.095, .117] &  \\
0.3 & positive-only SFT & .867 [.856, .879] & $-$.038 [$-$.046, $-$.030] & .043 [.034, .054] & .097 [.087, .109] &  \\
\bottomrule
\end{tabular}
\end{table}

\begin{table}[H]
\centering
\footnotesize
\setlength{\tabcolsep}{3pt}
\caption{Noise-rate sweep, N 5000, seed 0: calibration to the channel and selection. ``Final step'' is step 1407, the end of the budget. Criterion growth is the selection criterion's deterioration at the final step minus at the selected step, defined below the table.}
\label{tab:A4b}
\begin{tabular}{llrrrrrrr}
\toprule
f & learner & channel-scale ECE & reference & mean top-1 & channel line & selected step & final-step accuracy & criterion growth \\
\midrule
0 & CAPS-Brier &  &  &  &  & 300 & .933 & .059 \\
0.1 & CAPS-Brier & .038 [.035, .046] & .021 & .870 & .838 & 450 & .892 & .114 \\
0.2 & CAPS-Brier & .058 [.052, .063] & .014 & .743 & .744 & 250 & .813 & .284 \\
0.3 & CAPS-Brier & .069 [.065, .073] & .009 & .674 & .655 & 250 & .734 & .757 \\
0 & positive-only SFT &  &  &  &  & 200 &  & .268 \\
0.1 & positive-only SFT & .124 [.118, .133] & .063 & .935 & .812 & 100 &  & 1.244 \\
0.2 & positive-only SFT & .196 [.190, .203] & .047 & .915 & .720 & 250 &  & 2.216 \\
0.3 & positive-only SFT & .251 [.247, .256] & .031 & .896 & .647 & 100 &  & 2.916 \\
\bottomrule
\end{tabular}
\end{table}

Channel-scale ECE scores the stored top-1 probability against the channel's outcome rate, $f + (1 - 2f) \times \mathrm{correct}$. The reference is the clean run's top-1 probability mapped through $f + (1 - 2f)\,p$ and scored the same way, a channel-calibrated copy of the clean run. The channel line is $f + (1 - 2f)$ times the run's accuracy. The best achievable value of the selection criterion \eqref{eq:select} is $-H(f)$ for flipped outcomes, with $H$ the binary entropy in nats, and 0 for clean ones. Its deterioration is that value minus the criterion, and criterion growth is the deterioration at the final step minus at the selected step. CAPS-Brier's excess channel-scale ECE over the reference grows with f: .017, .044 and .060. The soft-target diagnostic is a CAPS-Brier run at f 0.2 trained on the expected flipped outcome $f + (1 - 2f)\,r$, computed from the clean outcome. This is the infinite-draw limit. It needs the clean outcome, so it is a diagnostic, not a learner. It has no draw variance left, and its residual, channel-scale ECE .019 to .024 against a reference of .014 (project report R3, Section~4), bounds from above the cost of the misspecification noted in Section~\ref{sec:3}. The rest of the excess is single-draw variance.

\subsection{Scale: 1.7B against 0.6B at N 5000}\label{sec:A5}

\begin{landscape}
\begin{table}[H]
\centering
\footnotesize
\setlength{\tabcolsep}{2pt}
\caption{Scale: 1.7B against 0.6B at N 5000. The 1.7B columns are seed 0; CAPS-Brier's three-seed row and differences follow the tables.}
\label{tab:A5}
\begin{tabular}{llrrrrrrrr}
\toprule
learner & 1.7B accuracy & 1.7B ECE & 1.7B Brier & 1.7B NLL & 1.7B AURC & 1.7B Cov@.05 & 0.6B accuracy, seed 0 & 0.6B accuracy, seed mean (range) & 0.6B ECE, seed mean \\
\midrule
zero-shot & .848 [.836, .860] & .089 [.078, .100] & .242 & .660 & .0350 [.0301, .0401] & .731 [.699, .763] & .851 &  & .063 \\
temperature & .848 & .031 [.026, .044] & .222 & .519 & .0333 [.0285, .0382] & .750 [.719, .774] & .851 &  & .020 \\
positive-only SFT & .913 [.902, .923] & .067 [.058, .077] & .154 & .545 & .0177 [.0139, .0218] & .906 [.877, .933] & .905 & (seed 0 only) & .078 \\
CAPS-Brier & .949 [.941, .957] & .021 [.015, .028] & .079 & .188 & .0066 [.0049, .0084] & .998 [.981, 1.000] & .929 & .931 (.929, .935) & .025 \\
full-label SFT & .953 [.945, .960] & .023 [.018, .030] & .072 & .162 & .0053 [.0037, .0071] & 1.000 [.991, 1.000] & .947 & .946 (.943, .949) & .022 \\
\bottomrule
\end{tabular}
\end{table}

\begin{table}[H]
\centering
\footnotesize
\setlength{\tabcolsep}{3pt}
\caption{Scale: paired differences at 1.7B, N 5000. The seed-mean rows are a seed mean against one baseline seed, with the range of the per-seed differences below each.}
\label{tab:A5b}
\begin{tabular}{llrrrrr}
\toprule
1.7B paired difference & accuracy & ECE & Brier & NLL & AURC & Cov@.05 \\
\midrule
CAPS-Brier seed 0 minus positive-only SFT & +.036 [+.028, +.045] & $-$.046 [$-$.054, $-$.037] & $-$.075 [$-$.088, $-$.061] & $-$.357 [$-$.411, $-$.306] & $-$.0111 [$-$.0145, $-$.0083] & +.093 [+.065, +.115] \\
CAPS-Brier seed 0 minus full-label SFT & $-$.004 [$-$.010, +.002] & $-$.002 [$-$.008, +.004] & +.007 [+.001, +.013] & +.026 [+.013, +.040] & +.0013 [+.0004, +.0022] & $-$.002 [$-$.015, +.001] \\
CAPS-Brier seed mean minus positive-only SFT & +.034 [+.026, +.041] & $-$.051 [$-$.057, $-$.040] & $-$.070 [$-$.083, $-$.057] & $-$.353 [$-$.409, $-$.303] & $-$.0104 [$-$.0135, $-$.0078] & +.088 [+.062, +.108] \\
\quad per-seed range & (+.029, +.036) & ($-$.054, $-$.046) & ($-$.075, $-$.061) & ($-$.362, $-$.340) & ($-$.0111, $-$.0092) & (+.078, +.093) \\
CAPS-Brier seed mean minus full-label SFT & $-$.006 [$-$.011, $-$.002] & $-$.007 [$-$.011, +.000] & +.012 [+.006, +.017] & +.030 [+.018, +.043] & +.0021 [+.0012, +.0030] & $-$.007 [$-$.019, $-$.000] \\
\quad per-seed range & ($-$.011, $-$.004) & ($-$.010, $-$.002) & (+.007, +.021) & (+.020, +.042) & (+.0013, +.0032) & ($-$.017, $-$.002) \\
\bottomrule
\end{tabular}
\end{table}
\end{landscape}

CAPS-Brier at 1.7B over training seeds 0, 1 and 2, mean (range): accuracy .947 (.942, .949), ECE .016 (.013, .021), Brier .084 (.079, .093) and NLL .192 (.182, .205). AURC is .0073 (.0066, .0085) and Cov@.05 .993 (.983, .998). Every 1.7B CAPS-Brier seed is above every 0.6B one (.942 to .949 against .929 to .935).

\subsection{Full labels, 0.6B, three seeds}\label{sec:A6}

\begin{table}[H]
\centering
\setlength{\tabcolsep}{3pt}
\footnotesize
\caption{Full labels: unseen-schema ECE, 0.6B, three seeds. Differences are seed means against one SFT run, with the range of the per-seed differences in parentheses. The temperature of every arm, like SFT's, is fitted on the in-domain validation split.}
\label{tab:A6}
\begin{tabular}{llrr}
\toprule
run & unseen mean ECE (seeds 0, 1, 2) & minus SFT + T & with its own temperature, minus SFT + T \\
\midrule
SFT & .152 &  &  \\
SFT + T & .121 &  &  \\
continued cross-entropy & .171 (.185, .172, .156) & +.050 [+.044, +.056] (+.034, +.064) & +.002 [$-$.001, +.009] ($-$.006, +.011) \\
REINFORCE, outcome & .248 (.262, .244, .239) & +.127 [+.118, +.135] (+.118, +.141) & +.025 [+.018, +.034] (+.019, +.035) \\
REINFORCE, $r - p$ & .153 (.173, .155, .131) & +.031 [+.026, +.038] (+.010, +.051) & $-$.002 [$-$.006, +.004] ($-$.010, +.007) \\
CAPS-Brier & .159 (.173, .165, .139) & +.037 [+.032, +.043] (+.017, +.051) & +.001 [$-$.004, +.007] ($-$.005, +.006) \\
CAPS-log & .170 (.181, .172, .156) & +.048 [+.043, +.055] (+.035, +.060) & +.005 [+.000, +.011] ($-$.001, +.013) \\
\bottomrule
\end{tabular}
\end{table}

\begin{landscape}
\begin{table}[H]
\centering
\footnotesize
\setlength{\tabcolsep}{2.5pt}
\caption{Full labels: accuracy, 0.6B. SFT, the 0.6B SFT head, and the detached-Brier REINFORCE run are one run each; the arms are means of three seeds, with the seed range in parentheses. In-domain is every gold-dependent question of the in-domain test split, CLINC150 only: 11800 yes-or-no and choice questions, none of them score questions. The SST-5 and Yelp columns are the score questions of the separate SST-5 and Yelp test splits, 2210 and 1000. The last two columns are seed means against SFT's one run, with the range of the per-seed differences.}
\label{tab:A6b}
\begin{tabular}{>{\raggedright\arraybackslash}p{2.6cm}rrrrr}
\toprule
run & in-domain accuracy & SST-5 score accuracy & Yelp score accuracy & SST-5 score, minus SFT & Yelp score, minus SFT \\
\midrule
SFT & .956 [.952, .960] & .636 [.616, .655] & .505 [.474, .536] &  &  \\
REINFORCE, detached Brier reward & .948 [.943, .952] & .111 [.098, .124] & .044 [.032, .057] & $-$.525 [$-$.553, $-$.502] & $-$.461 [$-$.496, $-$.428] \\
continued cross-entropy & .957 [.953, .960] (.954, .959) & .626 [.608, .644] (.621, .631) & .499 [.471, .528] (.479, .515) & $-$.010 [$-$.019, $-$.002] ($-$.015, $-$.005) & $-$.006 [$-$.018, +.005] ($-$.026, +.010) \\
REINFORCE, outcome & .953 [.950, .957] (.948, .958) & .623 [.605, .640] (.609, .633) & .494 [.466, .522] (.479, .505) & $-$.014 [$-$.023, $-$.005] ($-$.027, $-$.004) & $-$.011 [$-$.023, +.002] ($-$.026, +.000) \\
REINFORCE, $r - p$ & .957 [.953, .960] (.952, .959) & .624 [.607, .642] (.617, .635) & .505 [.477, .533] (.486, .521) & $-$.012 [$-$.021, $-$.003] ($-$.019, $-$.001) & $-$.000 [$-$.011, +.011] ($-$.019, +.016) \\
CAPS-Brier & .957 [.953, .960] (.954, .959) & .622 [.604, .640] (.620, .626) & .506 [.477, .536] (.501, .510) & $-$.014 [$-$.023, $-$.005] ($-$.016, $-$.010) & +.001 [$-$.008, +.012] ($-$.004, +.005) \\
CAPS-log & .958 [.954, .961] (.954, .960) & .627 [.609, .646] (.620, .633) & .494 [.466, .524] (.481, .506) & $-$.009 [$-$.017, $-$.001] ($-$.016, $-$.003) & $-$.011 [$-$.021, $-$.001] ($-$.024, +.001) \\
\bottomrule
\end{tabular}
\end{table}
\end{landscape}

\begin{table}[H]
\centering
\small
\caption{The detached Brier reward simulated on the 0.6B SFT head's 42422 training questions, with one sampled group of 4 per question from $q = 0.9\,p + 0.1/K$. The columns give the share of questions, in percent, whose gold logit the resulting step lowers, raises or leaves unchanged. Yes-or-no questions have $K = 2$. The few unchanged questions in neither of the last two columns have non-zero advantages that cancel on the gold logit.}
\label{tab:A6c}
\begin{tabular}{lrrrrrr}
\toprule
questions & n & lowered & raised & unchanged & all samples identical & distinct, equal rewards \\
\midrule
all & 42422 & 24.0 & 1.8 & 74.2 & 63.7 & 10.4 \\
yes-or-no & 20399 & 0.0 & 0.0 & 100.0 & 78.4 & 21.6 \\
choice & 13482 & 30.8 & 0.9 & 68.4 & 67.9 & 0.0 \\
score & 8541 & 70.5 & 7.7 & 21.8 & 21.8 & 0.0 \\
\bottomrule
\end{tabular}
\end{table}

\subsection{Emotion, 0.6B}\label{sec:A7}

Positive-only SFT and full-label SFT have one seed on emotion. AURC is lower for CAPS-Brier than for positive-only SFT by .050 to .174, and at N 5000 its Cov@.05 is .595 against .032. Positive-only SFT trains on 206 to 2151 confirmed answers and is worse calibrated than zero-shot at every N (ECE .256 to .363 against .251). The zero-shot ECE of .251 here and the .207 of the 0.6B SFT head on the emotion split of R1 come from two constructions. This table's split has 2000 six-way choice questions. R1's emotion split scores 1000 six-way choice questions and 1000 yes-or-no questions together (.239 and .186 on each). Against full-label SFT, CAPS-Brier is 3.8 to 4.6 points below on seed 0. It closes 68, 79 and 88 percent of the accuracy gap between zero-shot (.540) and full-label SFT, computed from the stored accuracies. Its seeds span accuracy .821 to .832 and ECE .031 to .070. CAPS-log at N 5000, one seed, is $-$.021 [$-$.030, $-$.012] below CAPS-Brier's seed mean in accuracy ($-$.026 to $-$.015 against each CAPS-Brier seed). It has higher Brier and NLL, so Brier stays the default. The temperature learner (T about 2.3) has the lowest ECE, .024 at N 5000, and leaves AURC at .294 against zero-shot's .295: it fixes the level, not the ranking. At N 5000, CAPS-Brier costs 1.9 to 2.6 points of in-domain accuracy across seeds and full-label SFT 3.4, both with in-domain ECE within .004 of zero-shot. The label audit finds a problem in .32 and .40 of a uniform sample, mostly a labelling convention that follows emotion words (.20), and no plainly wrong label. The ceiling is .93 or higher if the convention is learnable and near .80 if it is not (Appendix~\ref{sec:setup} gives the samples and how the shares give these ceilings), and full-label SFT and every CAPS-Brier seed are above .80. On the 500-record emotion subset, gpt-4.1-mini scores .598. Paired over the same records, CAPS-Brier's seed mean is +.251 [+.209, +.295] above it (per seed, +.244 to +.260). The emotion labels follow a convention that has to be learned from the domain's labels, which the API model never sees, so this reference is not like for like.

\begin{table}[H]
\centering
\footnotesize
\setlength{\tabcolsep}{3pt}
\caption{Emotion, 0.6B: the N-curve (CAPS-Brier seed 0). Brackets are 95 percent record-bootstrap intervals; AURC and Cov@.05 rank by top-1 probability.}
\label{tab:A7}
\begin{tabular}{llrrrrr}
\toprule
N & metric & zero-shot & temperature & positive-only SFT & full-label SFT & CAPS-Brier \\
\midrule
500 & accuracy & .540 [.518, .562] & .540 & .578 [.555, .599] & .659 [.640, .681] & .621 [.601, .642] \\
2000 & accuracy & .540 & .540 & .589 [.567, .611] & .760 [.742, .779] & .715 [.695, .734] \\
5000 & accuracy & .540 & .540 & .615 [.592, .636] & .871 [.856, .884] & .832 [.816, .848] \\
500 & ECE & .251 [.232, .272] & .031 [.024, .056] & .363 [.342, .385] & .066 [.057, .091] & .087 [.075, .110] \\
2000 & ECE & .251 & .027 [.021, .053] & .324 [.304, .344] & .056 [.043, .075] & .066 [.049, .085] \\
5000 & ECE & .251 & .024 [.020, .051] & .256 [.237, .277] & .041 [.031, .055] & .070 [.059, .087] \\
500 & Brier & .687 [.657, .718] & .600 [.580, .620] & .764 [.724, .805] & .482 [.457, .505] & .534 [.509, .559] \\
2000 & Brier & .687 & .600 [.580, .620] & .708 [.671, .744] & .346 [.321, .369] & .422 [.399, .446] \\
5000 & Brier & .687 & .600 [.580, .619] & .628 [.593, .662] & .193 [.175, .213] & .264 [.241, .288] \\
500 & NLL & 1.675 [1.579, 1.759] & 1.251 [1.209, 1.292] & 3.147 [2.939, 3.327] & .980 [.927, 1.030] & 1.154 [1.094, 1.212] \\
2000 & NLL & 1.675 & 1.251 [1.209, 1.292] & 2.256 [2.111, 2.389] & .697 [.644, .750] & .877 [.830, .924] \\
5000 & NLL & 1.675 & 1.251 [1.210, 1.291] & 1.641 [1.532, 1.741] & .383 [.340, .428] & .599 [.541, .657] \\
500 & AURC & .295 [.272, .321] & .294 [.270, .320] & .262 [.236, .286] & .190 [.170, .209] & .212 [.193, .234] \\
2000 & AURC & .295 & .294 [.270, .320] & .254 [.231, .277] & .101 [.088, .115] & .148 [.130, .165] \\
5000 & AURC & .295 & .294 [.270, .320] & .233 [.211, .255] & .031 [.025, .037] & .058 [.048, .069] \\
500 & Cov@.05 & .025 [.007, .062] & .014 [.007, .055] & .002 [.001, .029] & .072 [.041, .118] & .068 [.024, .105] \\
2000 & Cov@.05 & .025 & .014 [.007, .055] & .030 [.011, .054] & .277 [.163, .389] & .107 [.065, .219] \\
5000 & Cov@.05 & .025 & .015 [.007, .055] & .032 [.005, .071] & .762 [.689, .813] & .595 [.470, .660] \\
\bottomrule
\end{tabular}
\end{table}

\begin{landscape}
\begin{table}[H]
\centering
\footnotesize
\setlength{\tabcolsep}{3pt}
\caption{Emotion, 0.6B: paired differences, CAPS-Brier seed 0 minus the other learner.}
\label{tab:A7b}
\begin{tabular}{llrrrrrr}
\toprule
N & vs & accuracy & ECE & Brier & NLL & AURC & Cov@.05 \\
\midrule
500 & positive-only SFT & +.043 [+.027, +.060] & $-$.276 [$-$.290, $-$.254] & $-$.230 [$-$.257, $-$.206] & $-$1.993 [$-$2.138, $-$1.841] & $-$.050 [$-$.064, $-$.036] & +.066 [+.019, +.101] \\
2000 & positive-only SFT & +.126 [+.109, +.144] & $-$.258 [$-$.282, $-$.232] & $-$.286 [$-$.313, $-$.261] & $-$1.379 [$-$1.482, $-$1.273] & $-$.106 [$-$.121, $-$.093] & +.077 [+.036, +.182] \\
5000 & positive-only SFT & +.217 [+.197, +.238] & $-$.186 [$-$.204, $-$.162] & $-$.364 [$-$.396, $-$.330] & $-$1.042 [$-$1.134, $-$.947] & $-$.174 [$-$.195, $-$.154] & +.564 [+.439, +.632] \\
500 & full-label SFT & $-$.038 [$-$.053, $-$.025] & +.021 [+.001, +.037] & +.053 [+.041, +.064] & +.174 [+.146, +.204] & +.022 [+.014, +.030] & $-$.005 [$-$.035, +.025] \\
2000 & full-label SFT & $-$.046 [$-$.060, $-$.033] & +.010 [$-$.009, +.027] & +.076 [+.066, +.087] & +.180 [+.154, +.206] & +.047 [+.039, +.054] & $-$.170 [$-$.242, $-$.030] \\
5000 & full-label SFT & $-$.039 [$-$.052, $-$.027] & +.029 [+.017, +.043] & +.070 [+.056, +.088] & +.216 [+.178, +.261] & +.027 [+.020, +.035] & $-$.167 [$-$.260, $-$.100] \\
\bottomrule
\end{tabular}
\end{table}

\begin{table}[H]
\centering
\footnotesize
\setlength{\tabcolsep}{3pt}
\caption{Emotion, 0.6B: CAPS-Brier at N 5000 over training seeds.}
\label{tab:A7c}
\begin{tabular}{llrrrrrrr}
\toprule
run & accuracy & ECE & Brier & NLL & AURC & Cov@.05 & minus seed 0, accuracy & minus seed 0, ECE \\
\midrule
seed 0 & .832 [.816, .848] & .070 [.059, .087] & .264 & .599 & .058 & .595 &  &  \\
seed 1 & .829 [.812, .845] & .031 [.024, .049] & .261 & .567 & .059 & .556 & $-$.003 [$-$.014, +.008] & $-$.039 [$-$.048, $-$.025] \\
seed 2 & .821 [.803, .837] & .038 [.028, .055] & .266 & .573 & .058 & .572 & $-$.011 [$-$.022, +.000] & $-$.032 [$-$.042, $-$.020] \\
mean of 3 training seeds (range) & .827 (.821, .832) & .046 (.031, .070) & .263 (.261, .266) & .580 (.567, .599) & .059 (.058, .059) & .575 (.556, .595) &  &  \\
\bottomrule
\end{tabular}
\end{table}

\begin{table}[H]
\centering
\footnotesize
\setlength{\tabcolsep}{3pt}
\caption{Emotion, 0.6B: CAPS-Brier seed mean minus the other learner, N 5000, a seed mean against one baseline seed, with the range of the per-seed differences below each row.}
\label{tab:A7d}
\begin{tabular}{llrrrrr}
\toprule
seed mean minus & accuracy & ECE & Brier & NLL & AURC & Cov@.05 \\
\midrule
positive-only SFT & +.213 [+.193, +.233] & $-$.210 [$-$.227, $-$.186] & $-$.364 [$-$.395, $-$.334] & $-$1.061 [$-$1.150, $-$.968] & $-$.174 [$-$.193, $-$.154] & +.543 [+.409, +.618] \\
\quad per-seed range & (+.206, +.217) & ($-$.225, $-$.186) & ($-$.367, $-$.362) & ($-$1.074, $-$1.042) & ($-$.174, $-$.173) & (+.525, +.564) \\
full-label SFT & $-$.043 [$-$.056, $-$.031] & +.006 [$-$.005, +.018] & +.070 [+.056, +.085] & +.197 [+.164, +.235] & +.028 [+.022, +.034] & $-$.188 [$-$.297, $-$.128] \\
\quad per-seed range & ($-$.050, $-$.039) & ($-$.010, +.029): unresolved & (+.068, +.072) & (+.184, +.216) & (+.027, +.028) & ($-$.206, $-$.167) \\
\bottomrule
\end{tabular}
\end{table}
\end{landscape}

\begin{table}[H]
\centering
\setlength{\tabcolsep}{2pt}
\footnotesize
\caption{Emotion, 0.6B: forgetting at N 5000, minus zero-shot (in-domain .956, unseen-intents .892).}
\label{tab:A7e}
\begin{tabular}{llrrr}
\toprule
run & in-domain accuracy & in-domain ECE & unseen-intents accuracy & unseen-intents ECE \\
\midrule
positive-only SFT & $-$.005 [$-$.007, $-$.002] & +.007 [+.004, +.009] & +.010 [+.006, +.013] & $-$.003 [$-$.007, +.000] \\
CAPS-Brier seed 0 & $-$.026 [$-$.031, $-$.023] & +.001 [$-$.004, +.004] & $-$.020 [$-$.027, $-$.013] & +.002 [$-$.005, +.009] \\
CAPS-Brier seed 1 & $-$.019 [$-$.023, $-$.015] & +.003 [$-$.002, +.009] & $-$.024 [$-$.031, $-$.017] & $-$.017 [$-$.025, $-$.011] \\
CAPS-Brier seed 2 & $-$.024 [$-$.027, $-$.020] & $-$.003 [$-$.007, +.004] & $-$.034 [$-$.042, $-$.026] & +.000 [$-$.008, +.007] \\
CAPS-log & $-$.022 [$-$.026, $-$.019] & $-$.001 [$-$.006, +.003] & $-$.011 [$-$.018, $-$.004] & $-$.013 [$-$.020, $-$.007] \\
full-label SFT & $-$.034 [$-$.038, $-$.029] & $-$.002 [$-$.007, +.001] & $-$.040 [$-$.048, $-$.032] & +.011 [+.003, +.018] \\
\bottomrule
\end{tabular}
\end{table}

\clearpage
\section{Setup and reproducibility details}\label{sec:setup}

\textbf{Model and training.} LoRA has r 16, alpha 32 and dropout 0.05 on the attention projections, and answers are read from the letter logits at each question's answer slot in one forward pass. Supervised training of the two SFT heads samples options per record, with negated yes-or-no phrasings. It runs 2 epochs over the CLINC150 and SST-5 training records, 1378 steps at effective batch 32 (micro-batch 8), with lr 2e-4, a linear warmup over the first 3 percent of the steps, linear decay to 0 and gradient clipping at 1.0. Every 200 steps it is validated on a fixed, seeded 1000-record subset of the validation split, and the step with the lowest validation NLL is kept. At both sizes that is the last step, 1378.

The full-label arms of Section~\ref{sec:4.2} start from the 0.6B SFT head and take 500 steps at effective batch 32 (micro-batch 8), with lr 5e-5, 20 warmup steps, linear decay to 0 and gradient clipping at 1.0. They sample $G = 4$ actions per question from $q = 0.9\,p + 0.1/K$ and clip the importance weight at 5. All of them except continued cross-entropy add $\beta\,\mathrm{KL}(p \,\|\, p_{\mathrm{SFT}})$ with $\beta = .02$. Every 100 steps each run is validated on a 1000-record subset of the validation split drawn from its seed. The evaluated adapter is the trained step with the lowest validation NLL; the detached-Brier REINFORCE run selected step 100. This selection reads gold labels, unlike the selection on logged outcomes \eqref{eq:select}.

In adaptation, all trained learners take the same number of steps, $\max(200, \lceil 10 \times 0.9 N / 32 \rceil)$, at effective batch 32 (micro-batch 8) and lr 5e-5, with no KL term. Bootstrap intervals use 1000 resamples over records, paired across runs.

\textbf{Compute.} Every training run fits one Kaggle session on one NVIDIA T4 (\texttt{docs/KAGGLE.md}). The paper's 66 training runs, the run directories with a \texttt{train\_summary.json} other than two pipeline checks, took 59.0 GPU hours of training on a T4 in total, summed from the wall time each summary records. Evaluation is not included.

\textbf{Data licences.} As stated on 11 November 2026 on the Hugging Face card of each pinned dataset: CLINC150 (\texttt{clinc/clinc\_oos}), CC BY 3.0, as in its source repository's licence file; SST-5 (\texttt{SetFit/sst5}), no licence stated at source; AG News (\texttt{fancyzhx/ag\_news}), no licence stated at source, the card's licence field reading unknown; Yelp (\texttt{Yelp/yelp\_review\_full}), other, with a pointer to the Yelp dataset agreement; emotion (\texttt{dair-ai/emotion}), other, for educational and research purposes only; Banking77 (\texttt{legacy-datasets/banking77}), CC BY 4.0.

\textbf{Group baselines.} No full-label run divides the advantage by the group standard deviation: the setting is off in the full-label configuration and in the stored configuration of all 31 full-label runs (Appendix~\ref{sec:B}, S4). The full-label group baseline therefore subtracts the group mean only; the logged REINFORCE arm subtracts the micro-batch mean only.

\textbf{Learners.}
\begin{itemize}
  \item Temperature fitted to the logged outcomes: post-hoc and global, and it changes no argmax.
  \item Positive-only SFT: cross-entropy on the logged action of every interaction whose outcome is 1; outcome-0 interactions are skipped. A common practice.
  \item Full-label SFT: cross-entropy on gold for every logged interaction, gold read from the data file, never from the log. The upper bound, at one label per interaction.
  \item CAPS-Brier and CAPS-log: $(r - p_a)^2$ or $-(r \log p_a + (1 - r) \log(1 - p_a))$ on the logged action's probability for every interaction, positives and negatives, with no weights.
\end{itemize}

\textbf{Leak and duplicate gates.} Before any GPU run, the data must pass two CPU checks:
\begin{itemize}
  \item state-free leak probes: classifiers on phrasing, question text and structure, which must not beat chance by more than a fixed margin and a shuffled-target null;
  \item train-test duplicate rules.
\end{itemize}

The Banking77 log domain drops 30 messages that match the test split and 31 repeats.

\textbf{Off-policy forms}, all on the same log:
\begin{itemize}
  \item The logged REINFORCE arm with reward $r - p$: REINFORCE with reward $r - p_a$, detached, a micro-batch mean baseline and the importance weight $\min(p/q, 5)$, with $p$ the learner's current probability. In expectation it is the IPS-weighted CAPS-Brier loss, \eqref{eq:reinforce-ips} and Section~\ref{sec:3}, exactly while no weight is clipped. It clips 132 of 17800 weights over all training steps at N 2000 and 240 of 44916 at N 5000.
  \item IPS CAPS-log: CAPS-log with each interaction weighted by $w_i$ in \eqref{eq:ipsw}, $\min(1/q, 5)$ divided by its mean over the training prefix. This is importance-weighted regression; 216 of 1800 and 561 of 4500 prefix weights are clipped.
  \item IPS positive-only SFT: positive-only SFT with the same weights on the positives it keeps, of which 14 of 1332 and 25 of 3342 are clipped.
\end{itemize}

\textbf{Metrics.} For a test set of $n$ questions, let $p_{i,a}$ be the readout probability of option $a$ on question $i$, $g_i$ its gold option, $c_i$ its top-1 probability and $e_i \in \{0, 1\}$ its top-1 error. Accuracy is the mean of $1 - e_i$. The evaluation Brier score is multiclass, $\frac{1}{n} \sum_i \sum_a (p_{i,a} - 1[a = g_i])^2$ over all $K$ options, and NLL is $-\frac{1}{n} \sum_i \log p_{i,g_i}$, with $p_{i,g_i}$ floored at $10^{-12}$. Both score the whole readout against the gold option, unlike the chosen-action binary scores that CAPS trains on \eqref{eq:caps}. ECE sorts the questions by $c_i$ into 15 equal-width bins $B_1, \dots, B_{15}$ on $[0, 1]$:
\begin{equation}
  \label{eq:ece}
  \mathrm{ECE} = \sum_{b=1}^{15} \frac{|B_b|}{n} \Bigl| \frac{1}{|B_b|} \sum_{i \in B_b} (1 - e_i) - \frac{1}{|B_b|} \sum_{i \in B_b} c_i \Bigr|,
\end{equation}
with empty bins left out. Rank the questions by $c_i$, highest first, and let $\mathrm{risk}(m)$ be the error rate of the first $m$. AURC is the mean error of the kept answers over all coverages, and Cov@eps the largest share kept with error at most eps:
\begin{equation}
  \label{eq:aurc}
  \mathrm{AURC} = \frac{1}{n} \sum_{m=1}^{n} \mathrm{risk}(m),
\end{equation}
\begin{equation}
  \label{eq:cov}
  \mathrm{Cov}@\epsilon = \max \Bigl\{ \frac{m}{n} : \mathrm{risk}(m) \le \epsilon \Bigr\},
\end{equation}
or 0 when no $m$ qualifies. Ties in top-1 are kept or dropped together. In \eqref{eq:cov} $m$ ends at the boundary of a tie group, and in \eqref{eq:aurc} a tie group contributes its mean over every order of its members. Cov@eps is computed for eps in \{.05, .10, .20\}. On Banking77 at 0.6B every learner reaches Cov@.20 = 1 at every N. Cov@.10 is 1 for every trained learner at N 5000, and at N 2000 for all but positive-only SFT and IPS positive-only SFT (.985 and .987) (Appendix~\ref{sec:A1}), so the tables show Cov@.05. The same metrics ranked by the response's confidence field, $1 - H/\ln K$ with $H$ the entropy of the readout, are also stored, under the keys named in Appendix~\ref{sec:B}.

\textbf{Label audits.} Two language-model judges, no human, audited the labels: Claude (Fable 5.1 and Opus 5.5, via Claude Code) and GPT-Astra. The Banking77 pass was committed on 4 October 2026 and the emotion pass on 15 October 2026. Exact model identifiers were not recorded. Each judge gave every sampled test question one verdict: label correct, ambiguous, label wrong, or convention, a label that follows an annotation rule the option descriptions do not state. A label problem is any verdict other than label correct.

On Banking77 the judges read a uniform sample of 50 test questions and a shared-error sample of 100 questions that both full-label SFT and CAPS-Brier miss. In the uniform sample they find a label problem in .12 and .14 and a wrong label in .04, 2 of 50 for each judge. In the shared-error sample they find a problem in .66 and .62. The ceiling of about .96 is one minus the wrong-label share of the uniform sample; ambiguous and convention labels would lower it further. A uniform sample of 50 in-domain CLINC150 test questions shows no problem for either judge.

On emotion they read a uniform sample of 50 test questions and a shared-error sample of 100 that both full-label SFT and positive-only SFT miss. In the uniform sample they find a problem in .32 and .40, mostly convention, a label that follows an emotion word in the message (.20 for both judges), and no wrong label. In the shared-error sample they find a problem in .93 and .90. The two emotion ceilings of Appendix~\ref{sec:A7} come from the uniform sample. With no wrong label in 50, the upper end of the 95 percent Wilson interval for the wrong-label share is .07, so the ceiling against wrong labels alone is .93 or higher. A reader who goes only by the option descriptions would also miss the convention rows, .20 of the sample, so that reader's ceiling is near .80.

\textbf{Dataset quality and evaluation hygiene.} Our leak probes are leakage detection by exploratory models on non-target features \citep{kaufman2012leakage} against a permutation null \citep{ojala2010permutation}. Benchmark test sets contain label errors \citep{northcutt2021pervasive}. For Banking77, \citet{ying2022banking77errors} flag about 1,400 of 13,083 utterances, about 11 percent, in line with our audit's .12 and .14. The emotion dataset is distantly supervised from hashtags \citep{saravia2018carer}. For their own hashtag-labelled emotion data, \citet{abdulmageed2017emonet} report human validation at about 90 percent label accuracy. Our audit finds a problem in a third of a uniform sample.

\textbf{Commits of the runs:}
\begin{itemize}
  \item b5913ef: the v3 logs and the phase 3 runs;
  \item b6b9e85 and 409c983: the off-policy runs;
  \item 36e8ee1: the noise sweep;
  \item c9e29c6: the 1.7B runs;
  \item 344019b: the 1.7B CAPS-Brier seeds 1 and 2;
  \item c86d322: the emotion log and runs;
  \item b9c0257: AURC and Cov@eps added to every run and stage file, earlier values unchanged;
  \item fe7eac1, b834000 and 287bbb1: the paper's table and figure scripts, the main table, and the stage file regenerated with the positive-only SFT seed key.
\end{itemize}

\textbf{Tables and figures.} Every table and figure is regenerated on CPU from committed files, by the scripts listed in Appendix~\ref{sec:B}.

\clearpage
\section{Provenance}\label{sec:B}
\counterwithout{table}{subsection}
\counterwithin{table}{section}

Every number in the paper comes from a committed file in the repository at the commits listed in Appendix~\ref{sec:setup}, or from a file in the published run dataset \texttt{yuanphd/jevmark-runs} (Section~\ref{sec:7}): the logs and the per-record results behind every bootstrap interval. Table~\ref{tab:prov} gives, for each table (T, by its number), figure (F, by its number) and in-text result (S), the file and the keys within it; \texttt{<N>}, \texttt{<f>}, \texttt{<learner>}, \texttt{<arm>}, \texttt{<split>}, \texttt{<type>}, \texttt{<k>} and \texttt{<run>} stand for the log size, flip rate, learner, arm, evaluation split, question type, training seed and run directory. R1, R2 and R3 are the project reports \texttt{docs/RESULTS\_v1.md}, \texttt{RESULTS\_v2.md} and \texttt{RESULTS\_v3.md}, and the report column gives the section with the full analysis. \texttt{S:} is \texttt{runs/v3\_stage\_06b/metrics.json}, \texttt{N:} \texttt{runs/v3\_noise\_sweep\_06b/metrics.json}, \texttt{L:} \texttt{runs/v3\_stage\_17b/metrics.json} and \texttt{E:} \texttt{runs/v3\_stage\_emotion\_06b/metrics.json}. A per-seed range is the minimum and maximum of each seed's stored value minus the baseline's stored value.

Names. In every file and key, positive-only SFT is \texttt{positive\_sft}, full-label SFT \texttt{full\_sft}, CAPS-Brier \texttt{direct\_brier} and CAPS-log \texttt{direct\_log}. IPS CAPS-log is \texttt{ips\_log}, IPS positive-only SFT \texttt{ips\_positive}, and the logged REINFORCE arm with reward $r - p$ \texttt{outcome\_minus\_p\_log}. The 0.6B and 1.7B SFT heads are \texttt{sft\_06b} and \texttt{sft\_17b}, the adapter names on the Hub. The in-domain, unseen-intents and SST-5 test splits are \texttt{test\_indomain}, \texttt{test\_unseen\_intents} and \texttt{test\_sst5}, R1's emotion split is \texttt{test\_emotion}, and the Banking77 and emotion test splits of the adaptation experiments are \texttt{v3\_banking77\_test\_full} and \texttt{v3\_emotion\_test\_full}. Metrics ranked by the confidence field carry the suffix \texttt{\_entropy}. The full-label runs set \texttt{rlcd.normalize\_std: false}, and the logs at flip rates 0.1 and 0.3 are rebuilt by \texttt{scripts/\allowbreak{}derive\_\allowbreak{}flip\_\allowbreak{}log.py}. The full-label arms are \texttt{sft\_cont} (continued cross-entropy), \texttt{outcome}, \texttt{outcome\_minus\_p} (REINFORCE with $r - p$), \texttt{direct\_brier} and \texttt{direct\_log}, and the detached-Brier REINFORCE run is \texttt{rlcd\_06b\_brier\_s0}; the f 0.2 run of Table~\ref{tab:A1c} is \texttt{direct\_brier\_n5000\_s0\_noisy}, the second log draw \texttt{direct\_brier\_n5000\_s0\_log1}, and the released adapter \texttt{rlcd\_banking77\_06b} is \texttt{runs/v3\_06b\_direct\_brier\_n5000\_s0}.

{\footnotesize
\setlength{\tabcolsep}{4pt}
\begin{longtable}{@{}p{0.06\textwidth}>{\raggedright\arraybackslash}p{0.23\textwidth}>{\raggedright\arraybackslash}p{0.53\textwidth}>{\raggedright\arraybackslash}p{0.11\textwidth}@{}}
\caption{Provenance of every table, figure and in-text result.}\label{tab:prov}\\
\toprule
No. & Item & File: keys & Report \\
\midrule
\endfirsthead
\toprule
No. & Item & File: keys & Report \\
\midrule
\endhead
\bottomrule
\endfoot
T\ref{tab:main} & main-text summary at N 5000 & \texttt{S:}, \texttt{L:} and \texttt{E:} \texttt{n\_\allowbreak{}curve.\allowbreak{}5000.\allowbreak{}\{zero\_\allowbreak{}shot,temperature,positive\_\allowbreak{}sft,full\_\allowbreak{}sft\}.\allowbreak{}\{accuracy,ece,aurc\};\allowbreak{} seeds.\allowbreak{}training\_\allowbreak{}seeds.\allowbreak{}\{accuracy,ece,aurc\};\allowbreak{} seeds.\allowbreak{}full\_\allowbreak{}sft.\allowbreak{}training\_\allowbreak{}seeds.\allowbreak{}\{accuracy,ece,aurc\};\allowbreak{} in S: only, seeds.\allowbreak{}positive\_\allowbreak{}sft.\allowbreak{}training\_\allowbreak{}seeds.\allowbreak{}\{accuracy,ece,aurc\}; each read only where it holds more than one seed} (\texttt{scripts/\allowbreak{}main\_\allowbreak{}table.py}) & R3 \S\S3, 12, 13 \\
T\ref{tab:A1} & N-curve & \texttt{S:} \texttt{n\_\allowbreak{}curve.\allowbreak{}\textless{}N\textgreater{}.\allowbreak{}\textless{}learner\textgreater{} (each metric and its \_\allowbreak{}ci);\allowbreak{} n\_\allowbreak{}curve.\allowbreak{}\textless{}N\textgreater{}.\allowbreak{}\textless{}learner\textgreater{}.\allowbreak{}coverage\_\allowbreak{}at\_\allowbreak{}\{0.\allowbreak{}10,0.\allowbreak{}20\}} & R3 \S2 \\
T\ref{tab:A1b} & N-curve differences & \texttt{S:} \texttt{n\_\allowbreak{}curve.\allowbreak{}\textless{}N\textgreater{}.\allowbreak{}direct\_\allowbreak{}brier\_\allowbreak{}minus\_\allowbreak{}\{positive\_\allowbreak{}sft,full\_\allowbreak{}sft\}} & R3 \S2 \\
T\ref{tab:A1c} & forgetting, selected steps, .95 threshold & \texttt{S:} \texttt{forgetting\_\allowbreak{}table.\allowbreak{}\textless{}split\textgreater{}.\allowbreak{}learners.\allowbreak{}\textless{}run\textgreater{} (accuracy, ece, minus\_\allowbreak{}zero\_\allowbreak{}shot);\allowbreak{} selected\_\allowbreak{}steps.\allowbreak{}runs.\allowbreak{}v3\_\allowbreak{}06b\_\allowbreak{}\textless{}run\textgreater{} (selected\_\allowbreak{}step, selected\_\allowbreak{}epochs, last\_\allowbreak{}epochs);\allowbreak{} coverage.\allowbreak{}\textless{}run\textgreater{}.\allowbreak{}0.\allowbreak{}95} & R3 \S\S5, 6, 11 \\
T\ref{tab:A2} & training seeds & \texttt{S:} \texttt{seeds.\allowbreak{}training\_\allowbreak{}seeds;\allowbreak{} seeds.\allowbreak{}direct\_\allowbreak{}log.\allowbreak{}training\_\allowbreak{}seeds;\allowbreak{} seeds.\allowbreak{}full\_\allowbreak{}sft.\allowbreak{}training\_\allowbreak{}seeds;\allowbreak{} seeds.\allowbreak{}log\_\allowbreak{}seed\_\allowbreak{}1} & R3 \S\S3, 11 \\
T\ref{tab:A2b} & seed-mean differences & \texttt{S:} \texttt{seeds.\allowbreak{}training\_\allowbreak{}seed\_\allowbreak{}mean\_\allowbreak{}minus\_\allowbreak{}positive\_\allowbreak{}sft;\allowbreak{} seeds.\allowbreak{}training\_\allowbreak{}seed\_\allowbreak{}mean\_\allowbreak{}minus\_\allowbreak{}full\_\allowbreak{}sft\_\allowbreak{}seed\_\allowbreak{}mean;\allowbreak{} seeds.\allowbreak{}direct\_\allowbreak{}log\_\allowbreak{}seed\_\allowbreak{}mean\_\allowbreak{}minus\_\allowbreak{}direct\_\allowbreak{}brier\_\allowbreak{}seed\_\allowbreak{}mean;\allowbreak{} per-seed ranges from seeds.\allowbreak{}seed\_\allowbreak{}\textless{}k\textgreater{} and n\_\allowbreak{}curve.\allowbreak{}5000.\allowbreak{}positive\_\allowbreak{}sft} & R3 \S\S3, 11 \\
T\ref{tab:A3} & off-policy baselines & \texttt{S:} \texttt{n\_\allowbreak{}curve.\allowbreak{}\textless{}N\textgreater{}.\allowbreak{}\textless{}arm\textgreater{} (each metric and its \_\allowbreak{}ci);\allowbreak{} n\_\allowbreak{}curve.\allowbreak{}\textless{}N\textgreater{}.\allowbreak{}off\_\allowbreak{}policy.\allowbreak{}\textless{}arm\textgreater{}.\allowbreak{}minus\_\allowbreak{}direct\_\allowbreak{}brier;\allowbreak{} off\_\allowbreak{}policy.\allowbreak{}clipping.\allowbreak{}\textless{}arm\textgreater{}.\allowbreak{}\textless{}N\textgreater{}} & R3 \S11 \\
T\ref{tab:A4} & noise sweep, accuracy and calibration & \texttt{N:} \texttt{clean.\allowbreak{}\textless{}arm\textgreater{} (accuracy, ece);\allowbreak{} by\_\allowbreak{}rate.\allowbreak{}\textless{}f\textgreater{}.\allowbreak{}\textless{}arm\textgreater{} (accuracy, minus\_\allowbreak{}clean.\allowbreak{}accuracy, ece, channel.\allowbreak{}inverted\_\allowbreak{}top1\_\allowbreak{}ece);\allowbreak{} by\_\allowbreak{}rate.\allowbreak{}\textless{}f\textgreater{}.\allowbreak{}direct\_\allowbreak{}brier\_\allowbreak{}loss\_\allowbreak{}minus\_\allowbreak{}positive\_\allowbreak{}sft\_\allowbreak{}loss.\allowbreak{}accuracy} & R3 \S4 \\
T\ref{tab:A4b} & noise sweep, channel and selection & \texttt{N:} \texttt{by\_\allowbreak{}rate.\allowbreak{}\textless{}f\textgreater{}.\allowbreak{}\textless{}arm\textgreater{}.\allowbreak{}channel.\allowbreak{}channel\_\allowbreak{}scale;\allowbreak{} by\_\allowbreak{}rate.\allowbreak{}\textless{}f\textgreater{}.\allowbreak{}\textless{}arm\textgreater{}.\allowbreak{}\{selection,last\};\allowbreak{} clean.\allowbreak{}\textless{}arm\textgreater{}.\allowbreak{}\{selection,last\}} & R3 \S4 \\
T\ref{tab:A5} & scale & \texttt{L:} \texttt{n\_\allowbreak{}curve.\allowbreak{}5000.\allowbreak{}\textless{}learner\textgreater{}} \newline \texttt{S:} \texttt{n\_\allowbreak{}curve.\allowbreak{}5000.\allowbreak{}\textless{}learner\textgreater{};\allowbreak{} seeds.\allowbreak{}training\_\allowbreak{}seeds;\allowbreak{} seeds.\allowbreak{}full\_\allowbreak{}sft.\allowbreak{}training\_\allowbreak{}seeds} & R3 \S12 \\
T\ref{tab:A5b} & scale differences and the three-seed values after them & \texttt{L:} \texttt{n\_\allowbreak{}curve.\allowbreak{}5000.\allowbreak{}direct\_\allowbreak{}brier\_\allowbreak{}minus\_\allowbreak{}\{positive\_\allowbreak{}sft,full\_\allowbreak{}sft\};\allowbreak{} seeds.\allowbreak{}training\_\allowbreak{}seeds;\allowbreak{} seeds.\allowbreak{}training\_\allowbreak{}seed\_\allowbreak{}mean\_\allowbreak{}minus\_\allowbreak{}\{positive\_\allowbreak{}sft,full\_\allowbreak{}sft\};\allowbreak{} per-seed ranges from seeds.\allowbreak{}seed\_\allowbreak{}\textless{}k\textgreater{} and n\_\allowbreak{}curve.\allowbreak{}5000.\allowbreak{}\{positive\_\allowbreak{}sft,full\_\allowbreak{}sft\}} & R3 \S12 \\
T\ref{tab:A6} & full labels, unseen-schema ECE & \texttt{runs/\allowbreak{}rlcd\_\allowbreak{}stage2a\_\allowbreak{}06b/\allowbreak{}metrics.\allowbreak{}json}: \texttt{unseen\_\allowbreak{}schemas (arms.\allowbreak{}\textless{}arm\textgreater{}.\allowbreak{}mean\_\allowbreak{}ece\_\allowbreak{}by\_\allowbreak{}seed for the raw ranges);\allowbreak{} temperature\_\allowbreak{}ablation} \newline \texttt{runs/\allowbreak{}rlcd\_\allowbreak{}06b\_\allowbreak{}\textless{}arm\textgreater{}\_\allowbreak{}s\textless{}k\textgreater{}\_\allowbreak{}temp/\allowbreak{}metrics.\allowbreak{}json}: \texttt{splits.\allowbreak{}\textless{}split\textgreater{}.\allowbreak{}overall.\allowbreak{}ece, the four unseen schemas averaged, for the ranges with its own temperature} & R2 \S3 \\
T\ref{tab:A6b} & full labels, accuracy (both difference columns) & \texttt{runs/\allowbreak{}v2\_\allowbreak{}accuracy\_\allowbreak{}06b/\allowbreak{}metrics.\allowbreak{}json}: \texttt{splits.\allowbreak{}\textless{}split\textgreater{}.\allowbreak{}\textless{}type\textgreater{}.\allowbreak{}\{sft\_\allowbreak{}06b,reinforce\_\allowbreak{}brier\_\allowbreak{}s0\};\allowbreak{} splits.\allowbreak{}\textless{}split\textgreater{}.\allowbreak{}\textless{}type\textgreater{}.\allowbreak{}arms.\allowbreak{}\textless{}arm\textgreater{} (accuracy, accuracy\_\allowbreak{}ci, accuracy\_\allowbreak{}range, minus\_\allowbreak{}sft, minus\_\allowbreak{}sft\_\allowbreak{}ci, by\_\allowbreak{}seed)} & R2 \S\S2, 3 \\
T\ref{tab:A6c} & the detached reward, simulated & \texttt{runs/\allowbreak{}advantage\_\allowbreak{}simulation\_\allowbreak{}06b/\allowbreak{}metrics.\allowbreak{}json}: \texttt{rewards.\allowbreak{}brier.\allowbreak{}\textless{}type\textgreater{} (questions, gold\_\allowbreak{}logit\_\allowbreak{}down, gold\_\allowbreak{}logit\_\allowbreak{}up, gold\_\allowbreak{}logit\_\allowbreak{}zero)} & R2 \S2 \\
T\ref{tab:A7} & emotion N-curve & \texttt{E:} \texttt{n\_\allowbreak{}curve.\allowbreak{}\textless{}N\textgreater{}.\allowbreak{}\textless{}learner\textgreater{}} & R3 \S13 \\
T\ref{tab:A7b} & emotion differences & \texttt{E:} \texttt{n\_\allowbreak{}curve.\allowbreak{}\textless{}N\textgreater{}.\allowbreak{}direct\_\allowbreak{}brier\_\allowbreak{}minus\_\allowbreak{}\{positive\_\allowbreak{}sft,full\_\allowbreak{}sft\}} & R3 \S13 \\
T\ref{tab:A7c} & emotion seeds & \texttt{E:} \texttt{seeds.\allowbreak{}seed\_\allowbreak{}\textless{}k\textgreater{};\allowbreak{} seeds.\allowbreak{}minus\_\allowbreak{}seed\_\allowbreak{}0;\allowbreak{} seeds.\allowbreak{}training\_\allowbreak{}seeds} & R3 \S13 \\
T\ref{tab:A7d} & emotion seed-mean differences & \texttt{E:} \texttt{seeds.\allowbreak{}training\_\allowbreak{}seed\_\allowbreak{}mean\_\allowbreak{}minus\_\allowbreak{}\{positive\_\allowbreak{}sft,full\_\allowbreak{}sft\};\allowbreak{} per-seed ranges from seeds.\allowbreak{}seed\_\allowbreak{}\textless{}k\textgreater{}, n\_\allowbreak{}curve.\allowbreak{}5000.\allowbreak{}positive\_\allowbreak{}sft and seeds.\allowbreak{}full\_\allowbreak{}sft.\allowbreak{}seed\_\allowbreak{}0} & R3 \S13 \\
T\ref{tab:A7e} & emotion forgetting & \texttt{E:} \texttt{forgetting.\allowbreak{}\textless{}split\textgreater{}.\allowbreak{}minus\_\allowbreak{}zero\_\allowbreak{}shot} & R3 \S13 \\
F\ref{fig:setting} & the setting & schematic, no data & \\
F\ref{fig:v2_arm_ece} & full-label ECE & \texttt{runs/\allowbreak{}rlcd\_\allowbreak{}stage2a\_\allowbreak{}06b/\allowbreak{}metrics.\allowbreak{}json}: \texttt{unseen\_\allowbreak{}schemas;\allowbreak{} temperature\_\allowbreak{}ablation} \newline \texttt{runs/\allowbreak{}rlcd\_\allowbreak{}06b\_\allowbreak{}\textless{}arm\textgreater{}\_\allowbreak{}s\textless{}k\textgreater{}\_\allowbreak{}temp/\allowbreak{}metrics.\allowbreak{}json}: \texttt{per-seed values after temperature} & R2 \S3 \\
F\ref{fig:v3_n_curve} & N-curve & \texttt{S:} \texttt{n\_\allowbreak{}curve.\allowbreak{}\textless{}N\textgreater{}.\allowbreak{}\textless{}learner\textgreater{};\allowbreak{} seeds.\allowbreak{}training\_\allowbreak{}seeds;\allowbreak{} seeds.\allowbreak{}full\_\allowbreak{}sft.\allowbreak{}training\_\allowbreak{}seeds} & R3 \S\S2, 3 \\
F\ref{fig:v4_noise_sweep} & noise sweep & \texttt{N:} \texttt{clean.\allowbreak{}\textless{}arm\textgreater{};\allowbreak{} by\_\allowbreak{}rate.\allowbreak{}\textless{}f\textgreater{}.\allowbreak{}\textless{}arm\textgreater{} (accuracy, minus\_\allowbreak{}clean, channel.\allowbreak{}channel\_\allowbreak{}scale, selection, last)} \newline \texttt{runs/\allowbreak{}\textless{}run\textgreater{}\_\allowbreak{}inverted/\allowbreak{}metrics.\allowbreak{}json}: \texttt{stored\_\allowbreak{}channel} & R3 \S4 \\
F\ref{fig:v4_scale} & scale & \texttt{S:} \texttt{n\_\allowbreak{}curve.\allowbreak{}5000.\allowbreak{}\textless{}learner\textgreater{};\allowbreak{} seeds.\allowbreak{}training\_\allowbreak{}seeds;\allowbreak{} seeds.\allowbreak{}full\_\allowbreak{}sft.\allowbreak{}training\_\allowbreak{}seeds;\allowbreak{} forgetting.\allowbreak{}test\_\allowbreak{}indomain.\allowbreak{}minus\_\allowbreak{}zero\_\allowbreak{}shot} \newline \texttt{L:} \texttt{the same keys} & R3 \S12 \\
F\ref{fig:v4_emotion_n_curve} & emotion N-curve & \texttt{E:} \texttt{n\_\allowbreak{}curve.\allowbreak{}\textless{}N\textgreater{}.\allowbreak{}\textless{}learner\textgreater{};\allowbreak{} n\_\allowbreak{}curve.\allowbreak{}5000.\allowbreak{}direct\_\allowbreak{}log;\allowbreak{} seeds.\allowbreak{}training\_\allowbreak{}seeds} & R3 \S13 \\
F\ref{fig:v4_offpolicy} & off-policy baselines & \texttt{S:} \texttt{n\_\allowbreak{}curve.\allowbreak{}\textless{}N\textgreater{}.\allowbreak{}\textless{}arm\textgreater{};\allowbreak{} seeds.\allowbreak{}training\_\allowbreak{}seeds;\allowbreak{} seeds.\allowbreak{}direct\_\allowbreak{}log.\allowbreak{}training\_\allowbreak{}seeds;\allowbreak{} seeds.\allowbreak{}full\_\allowbreak{}sft.\allowbreak{}training\_\allowbreak{}seeds} & R3 \S11 \\
S1 & Section~\ref{sec:1}: in-domain ECE .013 to .015, up to .22 on the unseen schemas & \texttt{runs/\allowbreak{}sft\_\allowbreak{}\{06b,17b\}/\allowbreak{}metrics.\allowbreak{}json}: \texttt{splits.\allowbreak{}\textless{}split\textgreater{}.\allowbreak{}overall.\allowbreak{}ece} & R1 \S2 \\
S2 & Section~\ref{sec:5}: the gap to full labels does not shrink with N & \texttt{S:} \texttt{n\_\allowbreak{}curve.\allowbreak{}\textless{}N\textgreater{}.\allowbreak{}direct\_\allowbreak{}brier\_\allowbreak{}minus\_\allowbreak{}full\_\allowbreak{}sft} & R3 \S8 \\
S3 & Section~\ref{sec:3}: negatives off the top choice, the uniform term's responsibility & \texttt{runs/\allowbreak{}v3\_\allowbreak{}log\_\allowbreak{}s0/\allowbreak{}metrics.\allowbreak{}json}: \texttt{negatives (on\_\allowbreak{}other, on\_\allowbreak{}other\_\allowbreak{}uniform\_\allowbreak{}share)} & R3 \S1; \texttt{docs/\allowbreak{}METHOD.md} \S2 \\
S4 & Sections~\ref{sec:1}, \ref{sec:3} and Appendix~\ref{sec:setup}: mean-only group baseline & \texttt{configs/\allowbreak{}rlcd\_\allowbreak{}06b.\allowbreak{}yaml;\allowbreak{} runs/\allowbreak{}rlcd\_\allowbreak{}06b\_\allowbreak{}*/\allowbreak{}config.\allowbreak{}yaml}: \texttt{rlcd.\allowbreak{}normalize\_\allowbreak{}std} & R2 \S2 \\
S5 & Appendix~\ref{sec:setup}: log-domain duplicates & \texttt{data/\allowbreak{}v3\_\allowbreak{}manifest.\allowbreak{}json} & R3 \S1 \\
S6 & Section~\ref{sec:4.1}: the logs (order, \texttt{other}, policy accuracy and outcome rate at both sizes) & \texttt{runs/\allowbreak{}v3\_\allowbreak{}log\_\allowbreak{}s0/\allowbreak{}metrics.\allowbreak{}json}: \texttt{sampling.\allowbreak{}order\_\allowbreak{}stream;\allowbreak{} chosen\_\allowbreak{}other\_\allowbreak{}rate;\allowbreak{} negatives.\allowbreak{}chosen\_\allowbreak{}other;\allowbreak{} accuracy;\allowbreak{} outcome\_\allowbreak{}rate} \newline \texttt{runs/\allowbreak{}v3\_\allowbreak{}log\_\allowbreak{}17b\_\allowbreak{}s0/\allowbreak{}metrics.\allowbreak{}json}: \texttt{accuracy;\allowbreak{} outcome\_\allowbreak{}rate;\allowbreak{} n} \newline \texttt{configs/\allowbreak{}data.\allowbreak{}yaml}: \texttt{unseen.\allowbreak{}banking77\_\allowbreak{}options} & R3 \S\S1, 6, 12 \\
S7 & Section~\ref{sec:4.1} and Appendix~\ref{sec:A1}: temperature against AURC and Cov@.05; the ECE interval example & \texttt{S:} \texttt{n\_\allowbreak{}curve.\allowbreak{}\{500,5000\}.\allowbreak{}\{zero\_\allowbreak{}shot,temperature\};\allowbreak{} n\_\allowbreak{}curve.\allowbreak{}5000.\allowbreak{}direct\_\allowbreak{}brier\_\allowbreak{}temp\_\allowbreak{}minus\_\allowbreak{}direct\_\allowbreak{}brier.\allowbreak{}aurc} & R3 \S2 \\
S8 & Section~\ref{sec:4.1}: score questions, equal steps, epochs, three seeds & \texttt{runs/\allowbreak{}v2\_\allowbreak{}accuracy\_\allowbreak{}06b/\allowbreak{}metrics.\allowbreak{}json}: \texttt{note} \newline \texttt{S:} \texttt{selected\_\allowbreak{}steps.\allowbreak{}runs.\allowbreak{}\textless{}run\textgreater{}.\allowbreak{}last\_\allowbreak{}epochs;\allowbreak{} selected\_\allowbreak{}steps.\allowbreak{}runs.\allowbreak{}v3\_\allowbreak{}06b\_\allowbreak{}positive\_\allowbreak{}sft\_\allowbreak{}n5000\_\allowbreak{}s0.\allowbreak{}n\_\allowbreak{}used;\allowbreak{} seeds} \newline \texttt{configs/\allowbreak{}v3\_\allowbreak{}06b.\allowbreak{}yaml}: \texttt{training.\allowbreak{}\{effective\_\allowbreak{}batch,micro\_\allowbreak{}batch\};\allowbreak{} v3.\allowbreak{}epochs} & R3 \S1 \\
S9 & Section~\ref{sec:3}: the IPS identity of \texttt{outcome\_minus\_p\_log} & \texttt{S:} \texttt{off\_\allowbreak{}policy.\allowbreak{}identity (claim, tiny\_\allowbreak{}model\_\allowbreak{}check.\allowbreak{}max\_\allowbreak{}rel\_\allowbreak{}difference\_\allowbreak{}reinforce\_\allowbreak{}vs\_\allowbreak{}constant\_\allowbreak{}times\_\allowbreak{}ips\_\allowbreak{}brier)} \newline \texttt{configs/\allowbreak{}v3\_\allowbreak{}06b.\allowbreak{}yaml}: \texttt{training.\allowbreak{}micro\_\allowbreak{}batch} & R3 \S11 \\
S10 & Section~\ref{sec:3}, Corollary~\ref{cor:detached}(iii) & \texttt{runs/\allowbreak{}advantage\_\allowbreak{}simulation\_\allowbreak{}06b/\allowbreak{}metrics.\allowbreak{}json}: \texttt{expected\_\allowbreak{}gold\_\allowbreak{}gradient (formula;\allowbreak{} rewards.\allowbreak{}\{brier,log\}.\allowbreak{}\{never\_\allowbreak{}positive,zero\_\allowbreak{}at\_\allowbreak{}k2,enumeration\_\allowbreak{}check\})} & R2 \S2 \\
S11 & Section~\ref{sec:4.2}: the broken reward and its simulation & \texttt{runs/\allowbreak{}v2\_\allowbreak{}accuracy\_\allowbreak{}06b/\allowbreak{}metrics.\allowbreak{}json}: \texttt{splits.\allowbreak{}\textless{}split\textgreater{}.\allowbreak{}\textless{}type\textgreater{}.\allowbreak{}\{sft\_\allowbreak{}06b,reinforce\_\allowbreak{}brier\_\allowbreak{}s0\}} \newline \texttt{runs/\allowbreak{}rlcd\_\allowbreak{}06b\_\allowbreak{}brier\_\allowbreak{}s0/\allowbreak{}config.\allowbreak{}yaml}: \texttt{rlcd.\allowbreak{}\{group\_\allowbreak{}size,epsilon,importance\_\allowbreak{}clip,normalize\_\allowbreak{}std,beta\}} \newline \texttt{runs/\allowbreak{}advantage\_\allowbreak{}simulation\_\allowbreak{}06b/\allowbreak{}metrics.\allowbreak{}json}: \texttt{settings;\allowbreak{} rewards.\allowbreak{}brier.\allowbreak{}overall} & R2 \S2 \\
S12 & Section~\ref{sec:4.2}: the full-label arms & \texttt{configs/\allowbreak{}rlcd\_\allowbreak{}06b.\allowbreak{}yaml}: \texttt{rlcd.\allowbreak{}\{group\_\allowbreak{}size,epsilon,importance\_\allowbreak{}clip,beta\}} \newline \texttt{runs/\allowbreak{}advantage\_\allowbreak{}simulation\_\allowbreak{}06b/\allowbreak{}metrics.\allowbreak{}json}: \texttt{expected\_\allowbreak{}gold\_\allowbreak{}gradient.\allowbreak{}\{max\_\allowbreak{}weight\_\allowbreak{}p\_\allowbreak{}over\_\allowbreak{}q,clip\_\allowbreak{}binds\}} \newline \texttt{runs/\allowbreak{}v2\_\allowbreak{}accuracy\_\allowbreak{}06b/\allowbreak{}metrics.\allowbreak{}json}: \texttt{splits.\allowbreak{}\textless{}split\textgreater{}.\allowbreak{}\textless{}type\textgreater{}.\allowbreak{}arms.\allowbreak{}\textless{}arm\textgreater{}} & R2 \S\S2, 3 \\
S13 & Section~\ref{sec:4.2}: no gain on full labels & \texttt{runs/\allowbreak{}rlcd\_\allowbreak{}stage2a\_\allowbreak{}06b/\allowbreak{}metrics.\allowbreak{}json}: \texttt{unseen\_\allowbreak{}schemas.\allowbreak{}mean\_\allowbreak{}ece.\allowbreak{}sft\_\allowbreak{}temp;\allowbreak{} unseen\_\allowbreak{}schemas.\allowbreak{}arms.\allowbreak{}\textless{}arm\textgreater{};\allowbreak{} temperature\_\allowbreak{}ablation.\allowbreak{}arms.\allowbreak{}\textless{}arm\textgreater{}} & R2 \S\S1, 3 \\
S14 & Section~\ref{sec:4.3} and Appendix~\ref{sec:A1}: the N-curve & \texttt{S:} \texttt{n\_\allowbreak{}curve.\allowbreak{}\textless{}N\textgreater{}.\allowbreak{}direct\_\allowbreak{}brier\_\allowbreak{}minus\_\allowbreak{}\{positive\_\allowbreak{}sft,full\_\allowbreak{}sft\};\allowbreak{} n\_\allowbreak{}curve.\allowbreak{}\textless{}N\textgreater{}.\allowbreak{}\textless{}learner\textgreater{}.\allowbreak{}accuracy;\allowbreak{} n\_\allowbreak{}curve.\allowbreak{}\textless{}N\textgreater{}.\allowbreak{}\textless{}learner\textgreater{}.\allowbreak{}ece} & R3 \S2 \\
S15 & Section~\ref{sec:4.3} and Appendix~\ref{sec:A2}: seeds & \texttt{S:} \texttt{seeds;\allowbreak{} direct\_\allowbreak{}brier\_\allowbreak{}temperatures} & R3 \S3 \\
S16 & Section~\ref{sec:4.3} and Appendices~\ref{sec:A3}, \ref{sec:setup}: off-policy baselines and clipping & \texttt{S:} \texttt{n\_\allowbreak{}curve.\allowbreak{}\textless{}N\textgreater{}.\allowbreak{}off\_\allowbreak{}policy.\allowbreak{}\textless{}arm\textgreater{}.\allowbreak{}minus\_\allowbreak{}\{direct\_\allowbreak{}brier,positive\_\allowbreak{}sft\};\allowbreak{} off\_\allowbreak{}policy.\allowbreak{}clipping.\allowbreak{}\textless{}arm\textgreater{}.\allowbreak{}\textless{}N\textgreater{} (prefix, trained\_\allowbreak{}on, prefix\_\allowbreak{}by\_\allowbreak{}outcome);\allowbreak{} selected\_\allowbreak{}steps.\allowbreak{}runs.\allowbreak{}\textless{}run\textgreater{}.\allowbreak{}selected\_\allowbreak{}epochs;\allowbreak{} seeds.\allowbreak{}direct\_\allowbreak{}log\_\allowbreak{}seed\_\allowbreak{}mean\_\allowbreak{}minus\_\allowbreak{}direct\_\allowbreak{}brier\_\allowbreak{}seed\_\allowbreak{}mean;\allowbreak{} seeds.\allowbreak{}\{training\_\allowbreak{}seeds,direct\_\allowbreak{}log.\allowbreak{}training\_\allowbreak{}seeds\}} \newline \texttt{runs/\allowbreak{}v3\_\allowbreak{}log\_\allowbreak{}s0/\allowbreak{}log.jsonl}: \texttt{propensity}, the negatives' weight floor & R3 \S11 \\
S17 & Section~\ref{sec:4.3} and Appendix~\ref{sec:A4}: noisy feedback & \texttt{N:} \texttt{by\_\allowbreak{}rate.\allowbreak{}\textless{}f\textgreater{}.\allowbreak{}direct\_\allowbreak{}brier.\allowbreak{}channel.\allowbreak{}channel\_\allowbreak{}scale.\allowbreak{}channel\_\allowbreak{}ece\_\allowbreak{}minus\_\allowbreak{}reference;\allowbreak{} by\_\allowbreak{}rate.\allowbreak{}\textless{}f\textgreater{}.\allowbreak{}\textless{}arm\textgreater{}.\allowbreak{}ece;\allowbreak{} by\_\allowbreak{}rate.\allowbreak{}\textless{}f\textgreater{}.\allowbreak{}direct\_\allowbreak{}brier.\allowbreak{}channel.\allowbreak{}inverted\_\allowbreak{}top1\_\allowbreak{}ece;\allowbreak{} by\_\allowbreak{}rate.\allowbreak{}\textless{}f\textgreater{}.\allowbreak{}amplification;\allowbreak{} clean.\allowbreak{}\textless{}arm\textgreater{}.\allowbreak{}ece;\allowbreak{} by\_\allowbreak{}rate.\allowbreak{}\textless{}f\textgreater{}.\allowbreak{}direct\_\allowbreak{}brier\_\allowbreak{}loss\_\allowbreak{}minus\_\allowbreak{}positive\_\allowbreak{}sft\_\allowbreak{}loss.\allowbreak{}accuracy;\allowbreak{} by\_\allowbreak{}rate.\allowbreak{}0.\allowbreak{}3.\allowbreak{}direct\_\allowbreak{}brier.\allowbreak{}last.\allowbreak{}accuracy;\allowbreak{} zero\_\allowbreak{}shot.\allowbreak{}accuracy;\allowbreak{} by\_\allowbreak{}rate.\allowbreak{}\textless{}f\textgreater{}.\allowbreak{}direct\_\allowbreak{}brier.\allowbreak{}selection.\allowbreak{}best\_\allowbreak{}step} \newline \texttt{runs/\allowbreak{}v3\_\allowbreak{}log\_\allowbreak{}s0\_\allowbreak{}flip0.\allowbreak{}3/\allowbreak{}metrics.\allowbreak{}json}: \texttt{outcome\_\allowbreak{}rate;\allowbreak{} flipped\_\allowbreak{}outcome\_\allowbreak{}rate;\allowbreak{} flip\_\allowbreak{}share} (mislabelled share of positives $= (o + s - \phi)/2$ subtracted from $s$, over $\phi$, for outcome rate $o$, flip share $s$ and flipped outcome rate $\phi$) & R3 \S4 \\
S18 & Section~\ref{sec:4.3} and Appendix~\ref{sec:A5}: scale & \texttt{L:} \texttt{seeds.\allowbreak{}training\_\allowbreak{}seeds;\allowbreak{} seeds.\allowbreak{}training\_\allowbreak{}seed\_\allowbreak{}mean\_\allowbreak{}minus\_\allowbreak{}\{positive\_\allowbreak{}sft,full\_\allowbreak{}sft\};\allowbreak{} n\_\allowbreak{}curve.\allowbreak{}5000.\allowbreak{}direct\_\allowbreak{}brier\_\allowbreak{}minus\_\allowbreak{}full\_\allowbreak{}sft;\allowbreak{} seeds.\allowbreak{}seed\_\allowbreak{}\textless{}k\textgreater{}} \newline \texttt{S:} \texttt{seeds.\allowbreak{}training\_\allowbreak{}seed\_\allowbreak{}mean\_\allowbreak{}minus\_\allowbreak{}full\_\allowbreak{}sft\_\allowbreak{}seed\_\allowbreak{}mean} & R3 \S12 \\
S19 & Section~\ref{sec:4.3}: forgetting and selective prediction & \texttt{S:} \texttt{forgetting\_\allowbreak{}table;\allowbreak{} coverage.\allowbreak{}\{positive\_\allowbreak{}sft,direct\_\allowbreak{}brier\}\_\allowbreak{}n5000\_\allowbreak{}s0.\allowbreak{}0.\allowbreak{}95;\allowbreak{} n\_\allowbreak{}curve.\allowbreak{}5000.\allowbreak{}\{direct\_\allowbreak{}brier,positive\_\allowbreak{}sft\}.\allowbreak{}coverage\_\allowbreak{}at\_\allowbreak{}0.\allowbreak{}05} & R3 \S\S5, 6 \\
S20 & Section~\ref{sec:4.3} and Appendix~\ref{sec:A7}: against an API model, paired & \texttt{runs/\allowbreak{}b2\_\allowbreak{}paired\_\allowbreak{}v3/\allowbreak{}metrics.\allowbreak{}json}: \texttt{domains.\allowbreak{}\{banking77,emotion\}.\allowbreak{}b2.\allowbreak{}accuracy\_\allowbreak{}all;\allowbreak{} domains.\allowbreak{}\textless{}domain\textgreater{}.\allowbreak{}seed\_\allowbreak{}means;\allowbreak{} domains.\allowbreak{}\textless{}domain\textgreater{}.\allowbreak{}learners.\allowbreak{}\textless{}learner\textgreater{}.\allowbreak{}minus\_\allowbreak{}b2} & R3 \S7 \\
S21 & Section~\ref{sec:4.4}: the emotion log & \texttt{runs/\allowbreak{}v3\_\allowbreak{}log\_\allowbreak{}emotion\_\allowbreak{}s0/\allowbreak{}metrics.\allowbreak{}json}: \texttt{accuracy;\allowbreak{} outcome\_\allowbreak{}rate;\allowbreak{} negatives.\allowbreak{}negatives;\allowbreak{} n;\allowbreak{} sampling.\allowbreak{}order\_\allowbreak{}stream} \newline \texttt{runs/\allowbreak{}v3\_\allowbreak{}log\_\allowbreak{}s0/\allowbreak{}metrics.\allowbreak{}json}: \texttt{outcome\_\allowbreak{}rate} & R3 \S13 \\
S22 & Section~\ref{sec:4.4} and Appendix~\ref{sec:A7}: emotion results & \texttt{E:} \texttt{n\_\allowbreak{}curve.\allowbreak{}\textless{}N\textgreater{}.\allowbreak{}direct\_\allowbreak{}brier\_\allowbreak{}minus\_\allowbreak{}\{positive\_\allowbreak{}sft,full\_\allowbreak{}sft\};\allowbreak{} n\_\allowbreak{}curve.\allowbreak{}\textless{}N\textgreater{}.\allowbreak{}\textless{}learner\textgreater{};\allowbreak{} seeds.\allowbreak{}training\_\allowbreak{}seeds;\allowbreak{} seeds.\allowbreak{}seed\_\allowbreak{}\textless{}k\textgreater{};\allowbreak{} seeds.\allowbreak{}full\_\allowbreak{}sft.\allowbreak{}seed\_\allowbreak{}0;\allowbreak{} seeds.\allowbreak{}training\_\allowbreak{}seed\_\allowbreak{}mean\_\allowbreak{}minus\_\allowbreak{}full\_\allowbreak{}sft;\allowbreak{} seeds.\allowbreak{}direct\_\allowbreak{}log;\allowbreak{} seeds.\allowbreak{}direct\_\allowbreak{}log\_\allowbreak{}seed\_\allowbreak{}mean\_\allowbreak{}minus\_\allowbreak{}direct\_\allowbreak{}brier\_\allowbreak{}seed\_\allowbreak{}mean;\allowbreak{} forgetting.\allowbreak{}test\_\allowbreak{}indomain.\allowbreak{}minus\_\allowbreak{}zero\_\allowbreak{}shot} \newline \texttt{runs/\allowbreak{}v3\_\allowbreak{}emotion\_\allowbreak{}06b\_\allowbreak{}positive\_\allowbreak{}sft\_\allowbreak{}n\textless{}N\textgreater{}\_\allowbreak{}s0/\allowbreak{}train\_\allowbreak{}summary.\allowbreak{}json}: \texttt{n\_\allowbreak{}used} \newline \texttt{docs/\allowbreak{}audit/\allowbreak{}emotion\_\allowbreak{}summary.\allowbreak{}json}: \texttt{groups} & R3 \S13 \\
S23 & Section~\ref{sec:4.4}: the comparison with Banking77 & \texttt{E:} \texttt{n\_\allowbreak{}curve.\allowbreak{}5000.\allowbreak{}direct\_\allowbreak{}brier\_\allowbreak{}minus\_\allowbreak{}full\_\allowbreak{}sft;\allowbreak{} seeds.\allowbreak{}training\_\allowbreak{}seed\_\allowbreak{}mean\_\allowbreak{}minus\_\allowbreak{}full\_\allowbreak{}sft} \newline \texttt{S:} \texttt{n\_\allowbreak{}curve.\allowbreak{}5000.\allowbreak{}direct\_\allowbreak{}brier\_\allowbreak{}minus\_\allowbreak{}full\_\allowbreak{}sft;\allowbreak{} seeds.\allowbreak{}training\_\allowbreak{}seed\_\allowbreak{}mean\_\allowbreak{}minus\_\allowbreak{}full\_\allowbreak{}sft\_\allowbreak{}seed\_\allowbreak{}mean} & R3 \S13 \\
S24 & Section~\ref{sec:2}: letter position & \texttt{runs/\allowbreak{}sft\_\allowbreak{}06b/\allowbreak{}metrics.\allowbreak{}json}: \texttt{splits.\allowbreak{}test\_\allowbreak{}indomain.\allowbreak{}letter\_\allowbreak{}bias.\allowbreak{}by\_\allowbreak{}position.\allowbreak{}\textless{}position\textgreater{}.\allowbreak{}accuracy} & R1 \S3 \\
S25 & Appendix~\ref{sec:A7}: the two emotion constructions & \texttt{runs/\allowbreak{}sft\_\allowbreak{}06b/\allowbreak{}metrics.\allowbreak{}json}: \texttt{splits.\allowbreak{}test\_\allowbreak{}emotion.\allowbreak{}\{overall,choice,noul\}.\allowbreak{}ece} \newline \texttt{E:} \texttt{n\_\allowbreak{}curve.\allowbreak{}\textless{}N\textgreater{}.\allowbreak{}zero\_\allowbreak{}shot.\allowbreak{}ece} & R1 \S2; R3 \S13 \\
S26 & Section~\ref{sec:5}: the selection comparison and the noisy run's in-domain ECE & \texttt{S:} \texttt{n\_\allowbreak{}curve.\allowbreak{}\{2000.\allowbreak{}full\_\allowbreak{}sft,5000.\allowbreak{}positive\_\allowbreak{}sft\}.\allowbreak{}accuracy;\allowbreak{} selected\_\allowbreak{}steps.\allowbreak{}runs.\allowbreak{}\textless{}run\textgreater{}.\allowbreak{}n\_\allowbreak{}used;\allowbreak{} forgetting\_\allowbreak{}table.\allowbreak{}test\_\allowbreak{}indomain.\allowbreak{}learners.\allowbreak{}direct\_\allowbreak{}brier\_\allowbreak{}n5000\_\allowbreak{}s0\_\allowbreak{}noisy.\allowbreak{}minus\_\allowbreak{}zero\_\allowbreak{}shot.\allowbreak{}ece} & R3 \S\S2, 5 \\
S27 & Section~\ref{sec:5} and Appendix~\ref{sec:setup}: label audits: samples, verdicts and ceilings & \texttt{docs/\allowbreak{}audit/\allowbreak{}summary.\allowbreak{}json;\allowbreak{} docs/\allowbreak{}audit/\allowbreak{}emotion\_\allowbreak{}summary.\allowbreak{}json}: \texttt{groups} (n, verdict shares, Wilson intervals); \texttt{docs/\allowbreak{}audit/\allowbreak{}SUMMARY.md}, \texttt{docs/\allowbreak{}audit/\allowbreak{}EMOTION\_\allowbreak{}SUMMARY.md}; the commit dates of \texttt{85ce118} and \texttt{4a0ae6d} (\texttt{git log -- docs/audit}) & \texttt{docs/audit} \\
S28 & Section~\ref{sec:5}: selection grid & \texttt{S:} \texttt{n\_\allowbreak{}curve.\allowbreak{}\textless{}N\textgreater{}.\allowbreak{}\textless{}arm\textgreater{};\allowbreak{} selected\_\allowbreak{}steps.\allowbreak{}runs.\allowbreak{}\textless{}run\textgreater{}.\allowbreak{}selected\_\allowbreak{}step} & R3 \S\S9, 11 \\
S29 & Section~\ref{sec:7}: the run-files dataset & \texttt{scripts/\allowbreak{}publish\_\allowbreak{}runs.\allowbreak{}py} (\texttt{REPO\_\allowbreak{}ID}, the files it places and the logs it leaves to \texttt{scripts/\allowbreak{}derive\_\allowbreak{}flip\_\allowbreak{}log.\allowbreak{}py}) & \\
S30 & Section~\ref{sec:2}: the timeline against OpenJev-RLCD & git tags \texttt{v1}, \texttt{v2}, \texttt{v3}, \texttt{v4-paper}: the commit date (\texttt{git log -1 --format=\%cd}) and the tag date of each & \\
S31 & Appendix~\ref{sec:setup}: supervised and full-label training, and their selection & \texttt{configs/\allowbreak{}sft\_\allowbreak{}\{06b,17b\}.\allowbreak{}yaml}, \texttt{configs/\allowbreak{}base.\allowbreak{}yaml}, \texttt{configs/\allowbreak{}rlcd\_\allowbreak{}06b.\allowbreak{}yaml}: \texttt{training;\allowbreak{} lora} \newline \texttt{runs/\allowbreak{}sft\_\allowbreak{}\{06b,17b\}/\allowbreak{}train\_\allowbreak{}summary.\allowbreak{}json}, \texttt{runs/\allowbreak{}rlcd\_\allowbreak{}06b\_\allowbreak{}brier\_\allowbreak{}s0/\allowbreak{}train\_\allowbreak{}summary.\allowbreak{}json}: \texttt{steps;\allowbreak{} best\_\allowbreak{}step} \newline \texttt{scripts/\allowbreak{}train\_\allowbreak{}sft.py}, \texttt{scripts/\allowbreak{}train\_\allowbreak{}rlcd.py}: selection by validation NLL & R1 \S1; R2 \S\S2, 3 \\
S32 & Section~\ref{sec:4.3} and Appendix~\ref{sec:A4}: clipping of backward-corrected values & \texttt{scripts/\allowbreak{}invert\_\allowbreak{}noisy.py}: \texttt{invert\_\allowbreak{}probs} and the inverted top-1 ECE, both clipped to $[0, 1]$ & R3 \S4 \\
S33 & Section~\ref{sec:4.1} and Appendix~\ref{sec:setup}: evaluation Brier and NLL & \texttt{jevmark/\allowbreak{}metrics.py}: \texttt{brier;\allowbreak{} nll;\allowbreak{} NLL\_\allowbreak{}FLOOR} & \\
S34 & Section~\ref{sec:4.2} and Table~\ref{tab:A6b}: the populations of the in-domain, SST-5 and Yelp columns & \texttt{runs/\allowbreak{}v2\_\allowbreak{}accuracy\_\allowbreak{}06b/\allowbreak{}metrics.\allowbreak{}json}: \texttt{note;\allowbreak{} splits.\allowbreak{}\textless{}split\textgreater{}.\allowbreak{}\textless{}type\textgreater{}.\allowbreak{}\{n,n\_\allowbreak{}records\}} & R2 \S2 \\
S35 & Appendix~\ref{sec:setup}: compute & \texttt{runs/\allowbreak{}*/\allowbreak{}train\_\allowbreak{}summary.\allowbreak{}json} (the 66 committed files, one per paper run): \texttt{wall\_\allowbreak{}clock\_\allowbreak{}seconds\_\allowbreak{}this\_\allowbreak{}session}, summed \newline \texttt{docs/\allowbreak{}KAGGLE.md}: one T4 per run & \\
S36 & Appendix~\ref{sec:setup}: data licences & \texttt{jevmark/\allowbreak{}data/\allowbreak{}sources.py}: the pinned dataset ids; each id's Hugging Face card, read 11 November 2026 (CLINC150 also its source repository's \texttt{LICENSE}) & \\
S37 & Section~\ref{sec:4.2}: the full-label protocol & \texttt{configs/\allowbreak{}rlcd\_\allowbreak{}06b.\allowbreak{}yaml}, \texttt{runs/\allowbreak{}rlcd\_\allowbreak{}06b\_\allowbreak{}*/\allowbreak{}config.\allowbreak{}yaml}: \texttt{training.\allowbreak{}\{steps,warmup\_\allowbreak{}steps,eval\_\allowbreak{}every,valid\_\allowbreak{}subset\}} \newline \texttt{runs/\allowbreak{}rlcd\_\allowbreak{}06b\_\allowbreak{}brier\_\allowbreak{}s0/\allowbreak{}train\_\allowbreak{}summary.\allowbreak{}json}: \texttt{best\_\allowbreak{}step} & R2 \S2 \\
\end{longtable}
}

{\sloppy \textbf{Scripts.} The tables are regenerated by \texttt{scripts/compare\_v3.py}, \texttt{scripts/compare\_noise.py}, \texttt{scripts/compare\_rlcd.py}, \texttt{scripts/paired\_deltas.py}, \texttt{scripts/log\_stats.py}, \texttt{scripts/v2\_accuracy\_table.py}, \texttt{scripts/simulate\_advantages.py}, \texttt{scripts/v3\_review.py} (the \texttt{off\_policy}, \texttt{forgetting\_table} and \texttt{selected\_steps} keys of \texttt{runs/v3\_stage\_06b/metrics.json}) \texttt{scripts/b2\_paired.py} and \texttt{scripts/main\_table.py} (Table~\ref{tab:main}, checked cell by cell against the appendix tables). \texttt{scripts/b2\_paired.py} reads the unpublished gpt-4.1-mini replies. Every figure is regenerated by \texttt{scripts/make\_figures.py}. The per-record files these scripts read are in the dataset \texttt{yuanphd/jevmark-runs}, placed there by \texttt{scripts/publish\_runs.py}. The training scripts (\texttt{scripts/train\_sft.py}, \texttt{scripts/train\_rlcd.py}) fit one Kaggle T4 session per run (\texttt{docs/KAGGLE.md}).\par}

\end{document}